\pdfoutput=1

\documentclass[pdflatex,sn-nature]{sn-jnl}

\usepackage{graphicx}%
\usepackage{multirow}%
\usepackage{amsmath,amssymb,amsfonts}%
\usepackage{amsthm}%
\usepackage{mathrsfs}%
\usepackage[title]{appendix}%
\usepackage{xcolor}%
\usepackage{textcomp}%
\usepackage{manyfoot}%
\usepackage{booktabs}%
\usepackage{algorithm}%
\usepackage{algorithmicx}%
\usepackage{algpseudocode}%
\usepackage{listings}%
\usepackage{subcaption}
\usepackage{xspace}

\newcommand{\yes}{\texttt{yes}\xspace}
\newcommand{\no}{\texttt{no}\xspace}
\newsavebox{\cirtab}

\theoremstyle{thmstyleone}%
\theoremstyle{thmstyletwo}%

\theoremstyle{thmstylethree}%

\begin{document}

\title[Single-Pass Decision Scores for Medical MLLMs]{Single-Pass Decision Scores Enable Rapid Classification with Adjustable Thresholds in Medical Multimodal Large Language Models}

\author*[1,2]{\fnm{Patrick} \sur{Wienholt}}\email{pwienholt@ukaachen.de}

\author[1,2]{\fnm{Frederik} \sur{Hauke}}

\author[1,2]{\fnm{Christiane} \sur{Kuhl}}
\author[3,4,5]{\fnm{Jakob Nikolas} \sur{Kather}}
\author[1,2]{\fnm{Sven} \sur{Nebelung}}
\author[1,2]{\fnm{Daniel} \sur{Truhn}}

\affil[1]{\orgdiv{Lab for Artificial Intelligence in Medicine, Department of Diagnostic and Interventional Radiology}, \orgname{University Hospital RWTH Aachen}, \orgaddress{\city{Aachen}, \country{Germany}}}

\affil[2]{\orgdiv{Department of Diagnostic and Interventional Radiology}, \orgname{University Hospital RWTH Aachen}, \orgaddress{\city{Aachen}, \country{Germany}}}

\affil[3]{\orgdiv{Department of Medical Oncology}, \orgname{National Center for Tumor Diseases (NCT), Heidelberg University Hospital}, \orgaddress{\city{Heidelberg}, \country{Germany}}}

\affil[4]{\orgdiv{Else Kroener Fresenius Center for Digital Health}, \orgname{Technical University Dresden}, \orgaddress{\city{Dresden}, \country{Germany}}}

\affil[5]{\orgdiv{Department of Medicine I, Faculty of Medicine and University Hospital Carl Gustav Carus}, \orgname{TUD Dresden University of Technology}, \orgaddress{\city{Dresden}, \country{Germany}}}

\abstract{Multimodal large language models (MLLMs) answer binary clinical questions with a \yes or \no token, leaving the decision threshold implicit and fixed. We derive a continuous score from a single forward pass by renormalizing the probabilities of the \yes and \no tokens, allowing the threshold to be adapted to clinical needs without retraining. We evaluated this approach with MedGemma and Gemma-3 (4B and 27B) on six medical imaging datasets. The default threshold was frequently misaligned with the task: for identifying normal chest radiographs, MedGemma-27B labeled every image as abnormal, whereas a validation-selected threshold yielded a sensitivity of 0.945 and a specificity of 0.696. Normalization achieved the highest AUROC on four of six datasets (CheXpert: 0.819 vs.\ 0.777 for raw and 0.743 for text-elicited probabilities), and a single decision required 53--79~ms on one GPU. This simple post-hoc method turns medical MLLMs into fast classifiers with adjustable operating points.\par\medskip\noindent\textbf{Code:} \url{https://github.com/TruhnLab/llm-threshold-calibration}}

\keywords{Large Language Models, Medical Imaging, Binary Classification, Decision Threshold, Deep Learning}

\maketitle

\section{Introduction}\label{sec:introduction}

The rapid evolution of deep learning has produced transformer-based large language models capable of performing complex tasks across various domains.
Multimodal large language models (MLLMs) extend these capabilities by jointly processing text and images, a feature of particular importance in medicine, where image-based data are central to diagnosis.
However, the development of MLLMs for medical diagnostics is still in its early stages~\cite{zhou2024preX22}, and significant challenges remain, particularly in achieving reliable confidence estimation and calibration~\cite{Wang2024Calibrated}.

Binary classification is a fundamental task in healthcare, essential for applications such as detecting the presence of a pathology, screening patient populations, or triaging urgent cases.
While classical machine learning models provide a tunable decision threshold to balance sensitivity and specificity, generative MLLMs typically lack this feature.
They answer binary questions by sampling the next token from a large vocabulary.
The model's implicit ``threshold'' is thus fixed by what causes one token to be sampled rather than another.

The raw probability of a specific answer token (e.g., \yes) is confounded by normalization over many unrelated tokens, limiting the utility of the resulting \yes/\no output in risk-sensitive clinical workflows where an interpretable confidence is crucial~\cite{Wang2024Calibrated}.
Prior work has addressed this gap from several angles.
Some approaches map classes to a small set of verbalizer tokens (e.g., \yes/\no), from which token-level probabilities are extracted for classification~\cite{schick2021TextClassificationX1,hegselmann23a}.
Others apply post-hoc calibration techniques; for instance, Giovannotti and Gammerman used an inductive Venn-Abers predictor to calibrate logits for binary label tokens, yielding well-calibrated probabilities that outperform simple temperature scaling~\cite{Giovannotti2024Calibrated}.
Complementary work on verbalized probability elicitation has highlighted calibration issues in models tuned with reinforcement learning from human feedback (RLHF) and introduced an ``invert softmax'' method with subsequent temperature scaling~\cite{Wang2024Calibrated}.
Beyond post-hoc adjustments, conformal prediction methods can provide coverage guarantees for binary decisions~\cite{cherian2024conformalPredictionX12}, and prompt-induced biases can be mitigated through contextual and batch calibration techniques that estimate and remove systematic offsets before a decision is made~\cite{zhao2021CalibrationX6,zhou2024BatchCalibrationX7}.
Reliability of binary answers can be improved by constraining the generation process.
Grammar- or schema-constrained decoding, for example, can restrict the answer space to valid tokens, eliminating invalid outputs with negligible overhead and without requiring fine-tuning~\cite{dong2024xgrammar}.
While these methods improve the calibration of output probabilities, they largely focus on general-purpose, text-based settings and leave the implicit decision threshold itself unaddressed. In medical, image-conditioned LLMs it remains unclear how far this default operating point is misaligned with clinically appropriate ones, and how simply it can be corrected.
Building on token-level verbalizers, we derive a binary confidence score from the relative next-token probabilities of the verbalizer tokens \yes and \no~\cite{hegselmann23a}.
Recently, dedicated decision models such as Jev have been proposed that return typed decisions with probabilities from a single pass instead of generating text, and are trained specifically to produce calibrated confidences~\cite{typesafe2026jev}.
Their designation as ``System One'' models refers to the fast, intuitive mode of thinking in dual-process theory~\cite{kahneman2011thinking,typesafe2026jev}.
Jev itself is proprietary and currently accepts text input only; first evaluations of decision models in medicine have accordingly focused on text-based and numerical tasks~\cite{madrid2026jevmedicine,du2026systemonemedicine}, and an image-based medical decision model inspired by Jev requires training a dedicated decision head and low-rank adapters on top of MedGemma~\cite{tang2026omnimedjev}.
At the same time, an early review of the first evaluations found that the typed readout itself has not yet shown an accuracy advantage over comparable label-probability readouts of ordinary LLMs~\cite{tang2026typedaudit}, which can themselves serve as Jev-style decision models~\cite{li2026llmasjev}.
Our approach follows a similar principle with existing open-weight multimodal LLMs: without any additional training, a single forward pass yields a probability for \yes, which can be thresholded directly.

Our contribution is threefold: (i) we adapt normalized verbalizer-token probabilities to medical MLLMs, (ii) we show that this produces a continuous, thresholdable score for binary decisions, and (iii) we demonstrate that normalization combined with post-hoc threshold adjustment uncovers latent discriminative signal, with the normalized score outperforming raw token probabilities and text-elicited estimates on most datasets.
Concretely, we adapt the approach to medical classification by using the first-step probabilities of \yes and \no as the basis for a continuous, tunable score.
We then normalize these probabilities over the two-token set.
This normalization isolates the model's belief in the affirmative class from the influence of all other tokens in the vocabulary.
Since only two tokens are considered, the resulting score corresponds directly to a tunable threshold.
A greedy decoding strategy is equivalent to a threshold of 0.5, and by varying this threshold, we can perform a receiver operating characteristic (ROC) analysis and compute the area under the ROC curve (AUROC).
This approach aligns with standard decision-analytic practices for selecting operating points in medicine~\cite{vickers2016netX25} without requiring any modification to the model itself.
Importantly, the clinically appropriate threshold is application-dependent: in screening settings one often prioritizes high sensitivity (accepting lower specificity) to minimize missed disease, whereas confirmatory settings may favor high specificity to avoid unnecessary follow-up or treatment.
For example, population screening for breast cancer on mammography typically targets high sensitivity, while deciding whether to initiate anticoagulation for suspected pulmonary embolism may warrant a higher-specificity operating point to limit harm from false positives.

\begin{figure}
    \centering
    \includegraphics[width=1\linewidth]{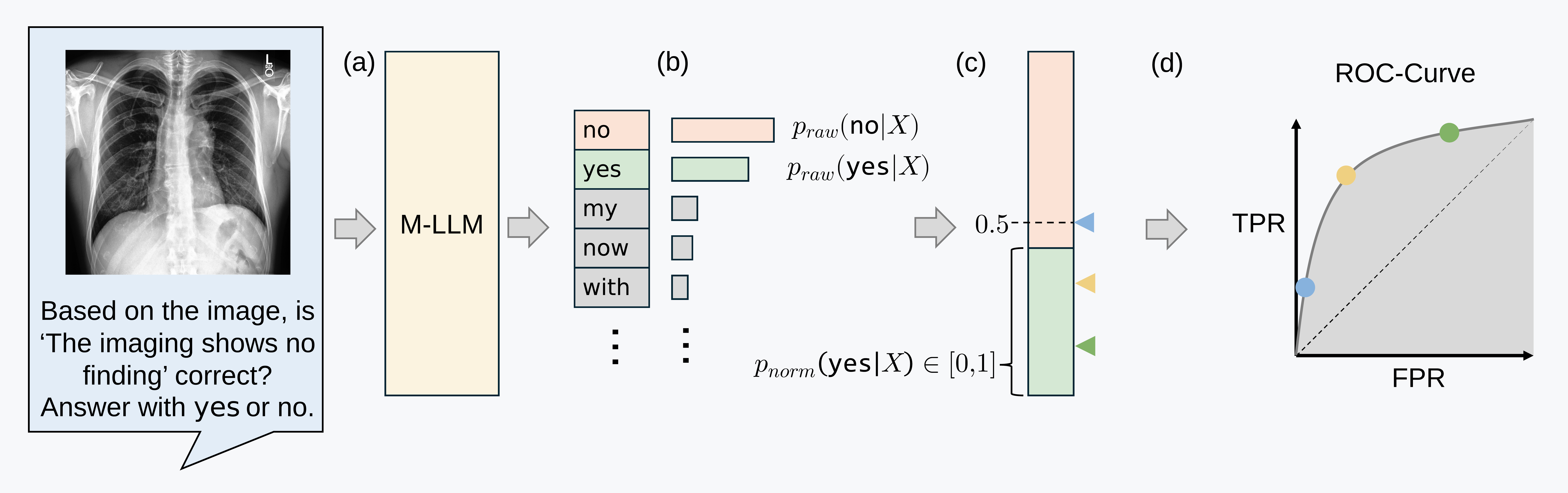}
    \caption{Workflow for threshold shifting in LLM-based binary classification. (a) An image-conditioned prompt is provided to the LLM. (b) From the first decoding step, the model assigns probabilities to all vocabulary tokens, including the verbalizers \yes and \no. (c) We normalize the probabilities over \{\texttt{yes}, \texttt{no}\} and vary the decision threshold, which has an implicit default of 0.5. (d) Sweeping the threshold yields different operating points on the ROC curve, trading true positive rate (TPR) against false positive rate (FPR). The optimal threshold was determined on a validation set disjoint from the test set.}
    \label{fig:workflow}
\end{figure}

\begin{figure}
    \centering
    \includegraphics[width=1\linewidth]{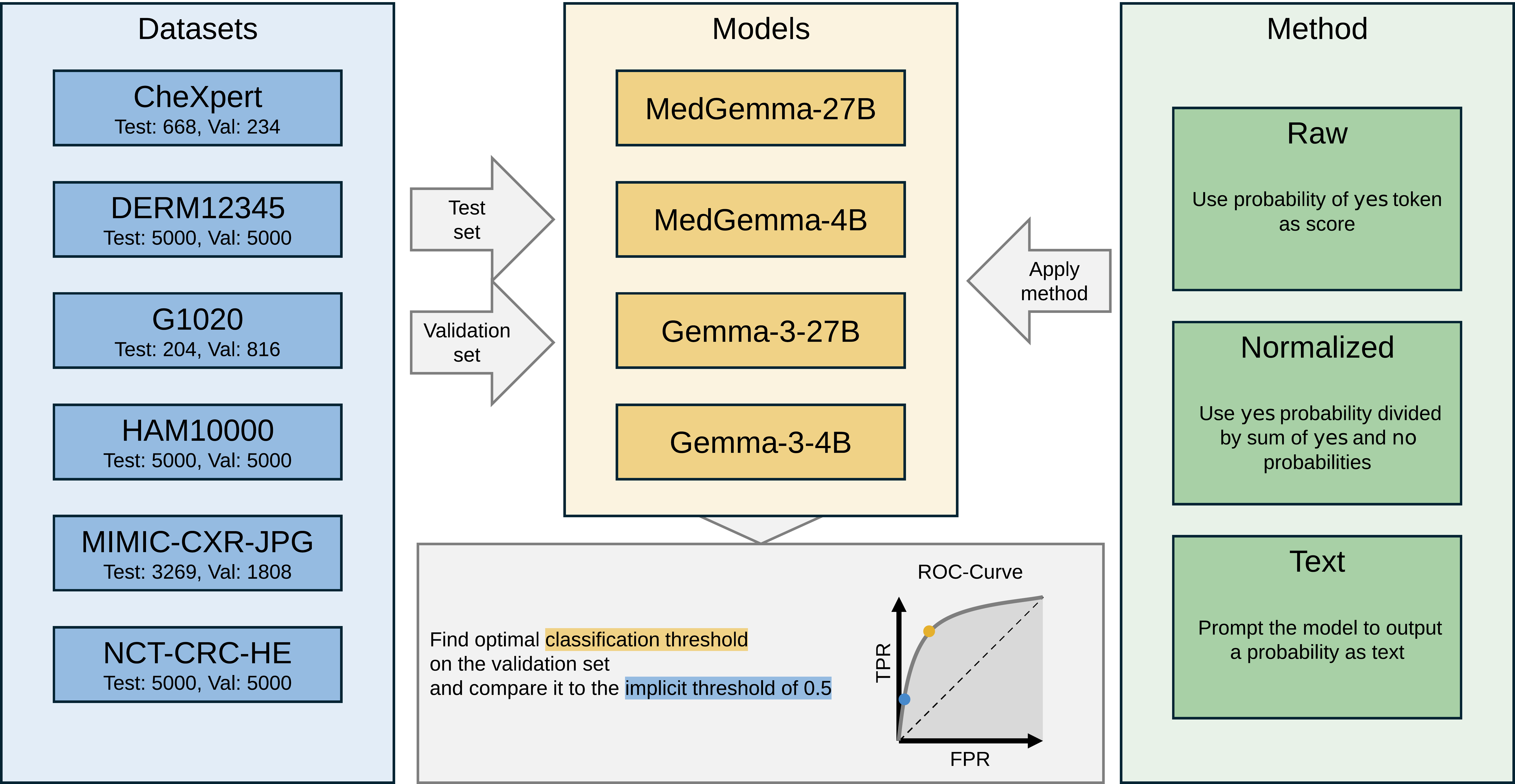}
    \caption{Overview of the experimental setup. We evaluated the threshold-shifting
method on six medical imaging datasets spanning chest radiography, dermatology,
ophthalmology, and histopathology. Notably, CheXpert was explicitly excluded from
the training data of the MedGemma models~\cite{sellergren2025medgemmatechnicalreport},
making it well suited for external validation. Each dataset was split into a
validation set, used exclusively to select the decision threshold, and a test set,
used only for the final performance evaluation. We compared four models, namely
Gemma-3 and MedGemma each in a 4B and a 27B variant, using three scoring methods:
the normalized \yes/\no token probability, the raw
\yes token probability, and a text-elicited probability generated directly as
output. The threshold optimized on the validation set was
compared against the model's implicit default threshold of 0.5 on the test set.}
    \label{fig:overview}
\end{figure}

Our approach is visualized in Fig.~\ref{fig:workflow}. We evaluate this approach across multiple medical imaging and diagnostic benchmarks using both medically fine-tuned models (MedGemma 4B/27B~\cite{sellergren2025medgemmatechnicalreport}) and their general-purpose counterparts (Gemma~3~4B/27B~\cite{gemmateam2025gemma3technicalreport}).
An overview of the full experimental setup, comprising the six datasets, four models, and three scoring methods, together with the disjoint validation/test split used for threshold selection and evaluation, is shown in Fig.~\ref{fig:overview}.
Our findings demonstrate that the implicit threshold used by generative models is often suboptimal.
Adjusting this threshold not only improves performance but also reveals diagnostic signals that would otherwise remain hidden.

\section{Results}\label{sec:results}

Because our method yields a continuous decision variable, we can freely choose the operating point on the ROC curve by setting the classification threshold, enabling task- and use-case-specific trade-offs between sensitivity and specificity.
In the following, we first report threshold-independent discrimination (AUROC) and then analyze how threshold selection affects clinically relevant operating characteristics.

\subsection{Overall performance (AUROC)}
First, we compared the binary classification ability of the three approaches, independent of a set threshold, using the AUROC.
Table~\ref{tab:medgemma27b_auroc_values} shows the average AUROC across all classes for the different datasets using MedGemma-27B.
All reported metrics stem from the independent test set, with decision thresholds derived solely from the validation split.
The values for the individual CheXpert classes are visualized in Figure \ref{fig:aurocMedGemma27BChexpert} and are provided in the supplementary material.
To assess statistical significance, we calculated pairwise differences and corresponding two-sided $p$-values (Table~\ref{tab:medgemma27b_auroc_diff}).

The normalization approach demonstrated a statistically significant improvement in mean AUROC for five of the six datasets compared to text-elicited probability estimates.
Similarly, compared to using raw token probabilities, normalization provided a statistically significant advantage on three of the six datasets. For the MIMIC and NCT-CRC-HE datasets, however, raw token probabilities yielded a small but statistically significant advantage.
For the G1020 dataset, no approach was significantly better than the others.

\begin{figure}
    \centering
    \includegraphics[width=\linewidth]{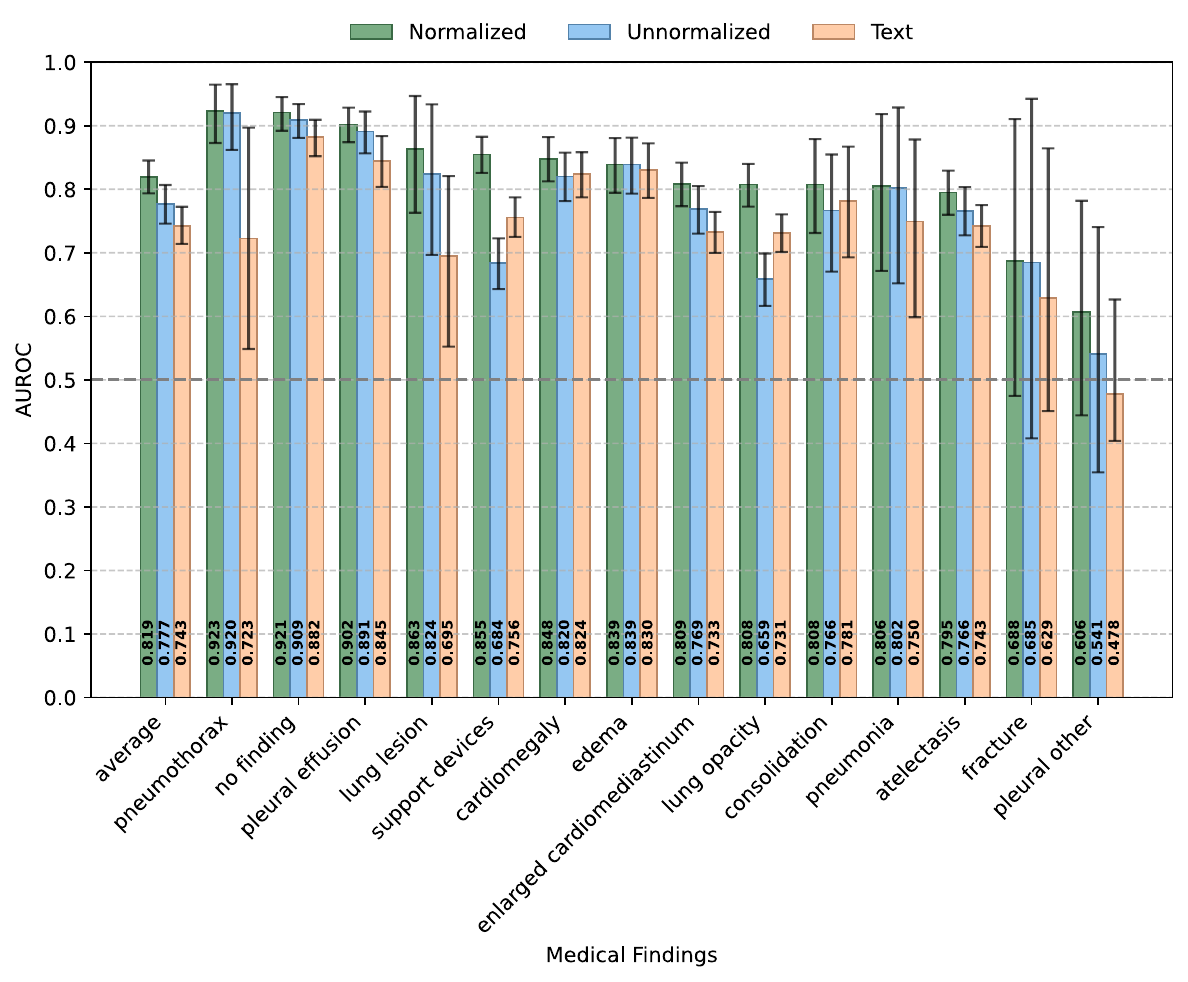}
    \caption{Class-specific AUROC values for the CheXpert dataset using MedGemma-27B. The three bars for each class represent normalized token probabilities, raw token probabilities (``Unnormalized'' in the legend), and text-elicited probability estimates (``Text''). Error bars indicate 95\% confidence intervals. The ``average'' bars show the mean across classes.}
    \label{fig:aurocMedGemma27BChexpert}
\end{figure}

\begin{table}[htbp]
\centering
\small
\caption{Average AUROC values computed over the test data with corresponding 95\% confidence intervals for MedGemma-27B.}
\label{tab:medgemma27b_auroc_values}
\begin{tabular}{lccc}
\toprule
Dataset & Normalized (95\% CI) & Raw (95\% CI) & Text-elicited (95\% CI) \\
\midrule
CheXpert & 0.819 [0.794, 0.845] & 0.777 [0.746, 0.807] & 0.743 [0.714, 0.772] \\
DERM12345 & 0.701 [0.682, 0.719] & 0.637 [0.613, 0.662] & 0.547 [0.530, 0.566] \\
G1020 & 0.494 [0.437, 0.551] & 0.489 [0.423, 0.555] & 0.487 [0.455, 0.519] \\
HAM10000 & 0.654 [0.642, 0.666] & 0.607 [0.592, 0.621] & 0.542 [0.531, 0.554] \\
MIMIC & 0.658 [0.647, 0.669] & 0.666 [0.655, 0.676] & 0.625 [0.616, 0.635] \\
NCT-CRC-HE & 0.783 [0.777, 0.788] & 0.796 [0.790, 0.801] & 0.642 [0.638, 0.647] \\
\bottomrule
\end{tabular}
\vspace{0.5ex}
{\footnotesize CI = 95\% confidence interval. ``Normalized'' refers to normalized token probabilities, ``Raw'' to the unnormalized $p_{\text{raw}}(\mathtt{yes}\mid X)$ score, and ``Text-elicited'' to probability estimates generated as text.}
\end{table}

\begin{table}[htbp]
\centering
\small
\caption{Pairwise comparison of average AUROC values computed over the test data  using MedGemma-27B from Table \ref{tab:medgemma27b_auroc_values}. The table shows the difference to the normalized approach. Statistically significant ($p < .0042$) values are in \textbf{bold}.}
\label{tab:medgemma27b_auroc_diff}
\begin{tabular}{lccccc}
\toprule
 & & \multicolumn{2}{c}{Raw} & \multicolumn{2}{c}{Text-elicited} \\
\cmidrule(lr){3-4}\cmidrule(lr){5-6}
Dataset & Normalized & $\Delta$ (95\% CI) & $p$ & $\Delta$ (95\% CI) & $p$ \\
\midrule
\multirow{2}{*}{CheXpert} & \multirow{2}{*}{0.819} & \textbf{-0.043} & \multirow{2}{*}{$\mathbf{< .001}$} & \textbf{-0.077} & \multirow{2}{*}{$\mathbf{< .001}$} \\
 &  & \textbf{[-0.057, -0.029]} &  & \textbf{[-0.095, -0.059]} &  \\
\addlinespace
\multirow{2}{*}{DERM12345} & \multirow{2}{*}{0.701} & \textbf{-0.064} & \multirow{2}{*}{$\mathbf{< .001}$} & \textbf{-0.154} & \multirow{2}{*}{$\mathbf{< .001}$} \\
 &  & \textbf{[-0.086, -0.041]} &  & \textbf{[-0.175, -0.133]} &  \\
\addlinespace
\multirow{2}{*}{G1020} & \multirow{2}{*}{0.494} & -0.005 & \multirow{2}{*}{$.873$} & -0.007 & \multirow{2}{*}{$.783$} \\
 &  & [-0.066, 0.054] &  & [-0.060, 0.046] &  \\
\addlinespace
\multirow{2}{*}{HAM10000} & \multirow{2}{*}{0.654} & \textbf{-0.047} & \multirow{2}{*}{$\mathbf{< .001}$} & \textbf{-0.111} & \multirow{2}{*}{$\mathbf{< .001}$} \\
 &  & \textbf{[-0.058, -0.037]} &  & \textbf{[-0.123, -0.100]} &  \\
\addlinespace
\multirow{2}{*}{MIMIC} & \multirow{2}{*}{0.658} & \textbf{0.008} & \multirow{2}{*}{$\mathbf{.001}$} & \textbf{-0.033} & \multirow{2}{*}{$\mathbf{< .001}$} \\
 &  & \textbf{[0.003, 0.013]} &  & \textbf{[-0.041, -0.024]} &  \\
\addlinespace
\multirow{2}{*}{NCT-CRC-HE} & \multirow{2}{*}{0.783} & \textbf{0.013} & \multirow{2}{*}{$\mathbf{< .001}$} & \textbf{-0.140} & \multirow{2}{*}{$\mathbf{< .001}$} \\
 &  & \textbf{[0.010, 0.016]} &  & \textbf{[-0.145, -0.135]} &  \\
\bottomrule
\end{tabular}
\vspace{0.5ex}
{\footnotesize Note: $\Delta = \text{AUROC}_{\text{comparator}} - \text{AUROC}_{\text{normalized}}$. CI = confidence interval.}
\end{table}


\subsection{Effect of the decision threshold}
The AUROC provides a threshold-independent measure of class separability. However, practical classification requires defining a specific decision threshold. We examined the threshold implicitly set by the LLM's default behavior of sampling the most likely token, which corresponds to a threshold of 0.5 in our normalized probability scheme.

\begin{figure*}[ht]
    \centering
    \begin{subfigure}[t]{0.3\textwidth}
        \centering
        \includegraphics[width=\linewidth]{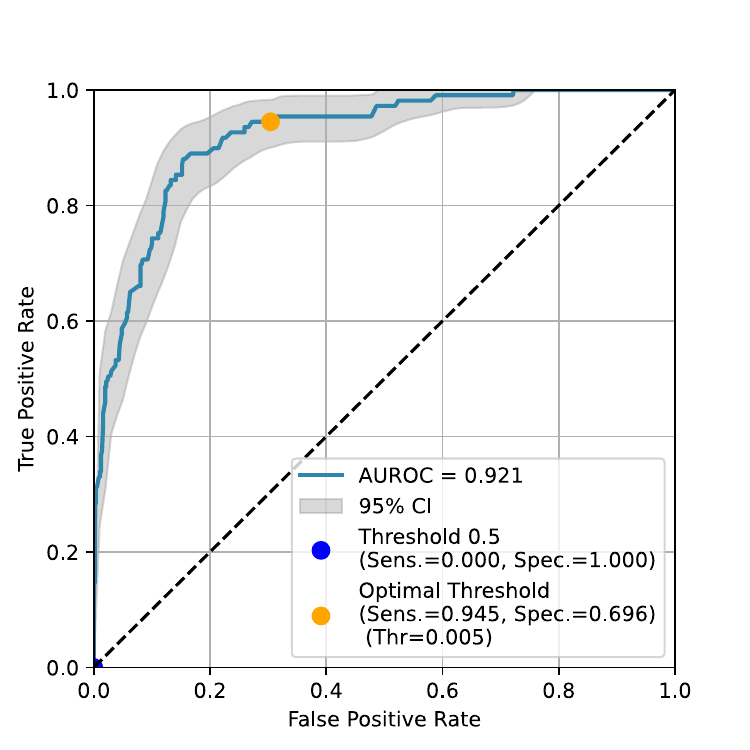}
        \caption{CheXpert - No Finding}
        \label{fig:noFindingROC_chexpert_27b}
    \end{subfigure}
    \hfill
    \begin{subfigure}[t]{0.3\textwidth}
        \centering
        \includegraphics[width=\linewidth]{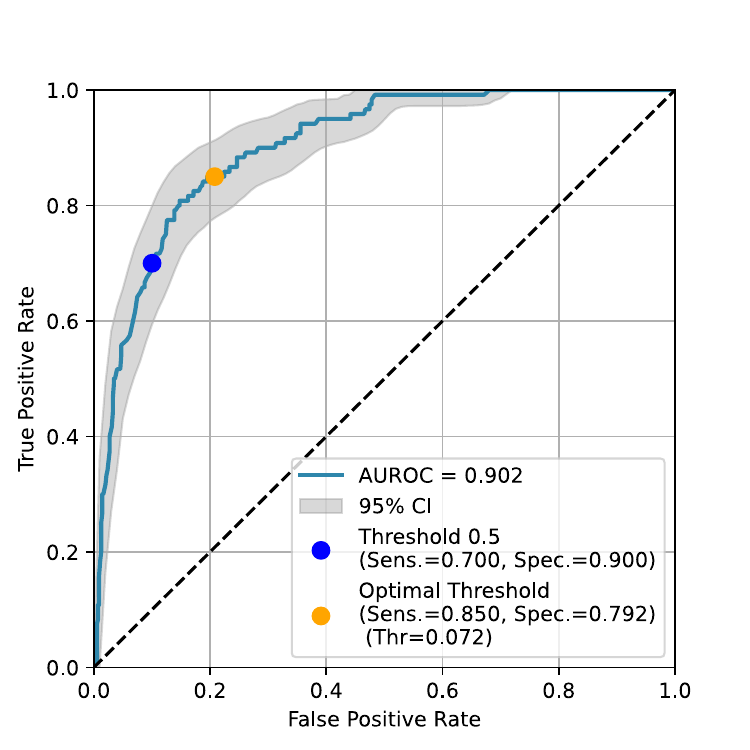}
        \caption{CheXpert - Pleural Effusion}
        \label{fig:pleuralEffusionROC_chexpert_27b}
    \end{subfigure}
    \hfill
    \begin{subfigure}[t]{0.3\textwidth}
        \centering
        \includegraphics[width=\linewidth]{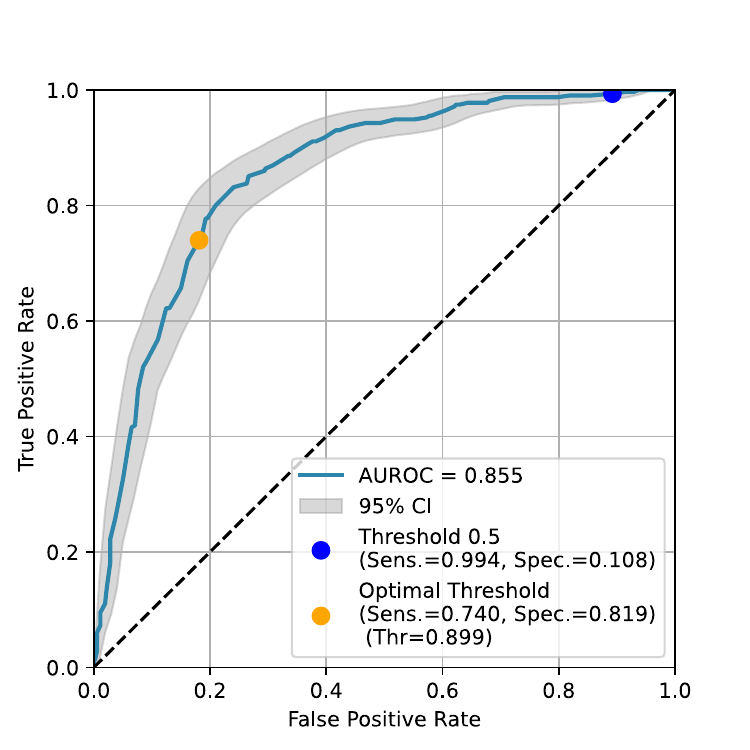}
        \caption{CheXpert - Support Devices}
        \label{fig:supportDevROC_chexpert_27b}
    \end{subfigure}
    
    \vspace{1em}
    
    \begin{subfigure}[t]{0.3\textwidth}
        \centering
        \includegraphics[width=\linewidth]{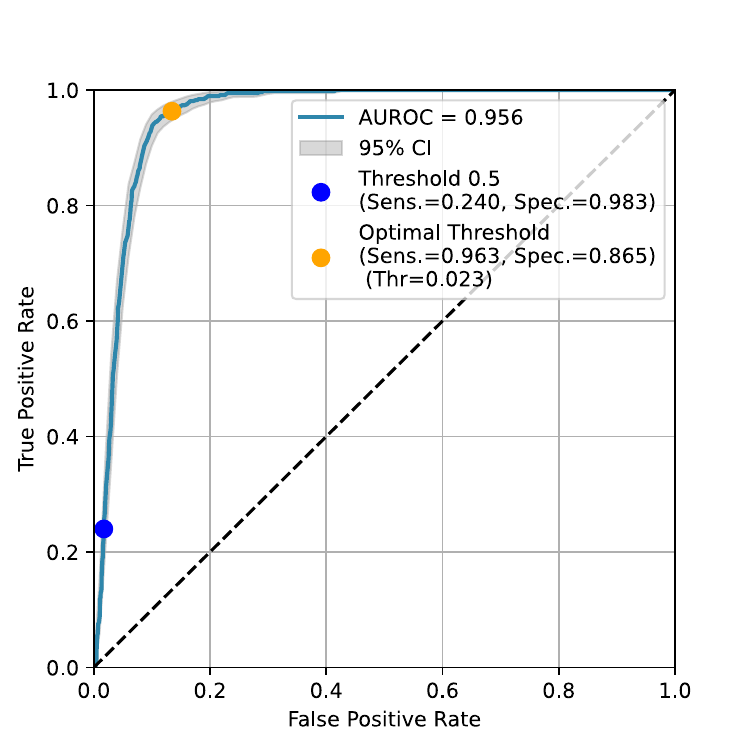}
        \caption{NCT-CRC-HE - Lymphocytes}
        \label{fig:lymphocytesROC_nct_27b}
    \end{subfigure}
    \hfill
    \begin{subfigure}[t]{0.3\textwidth}
        \centering
        \includegraphics[width=\linewidth]{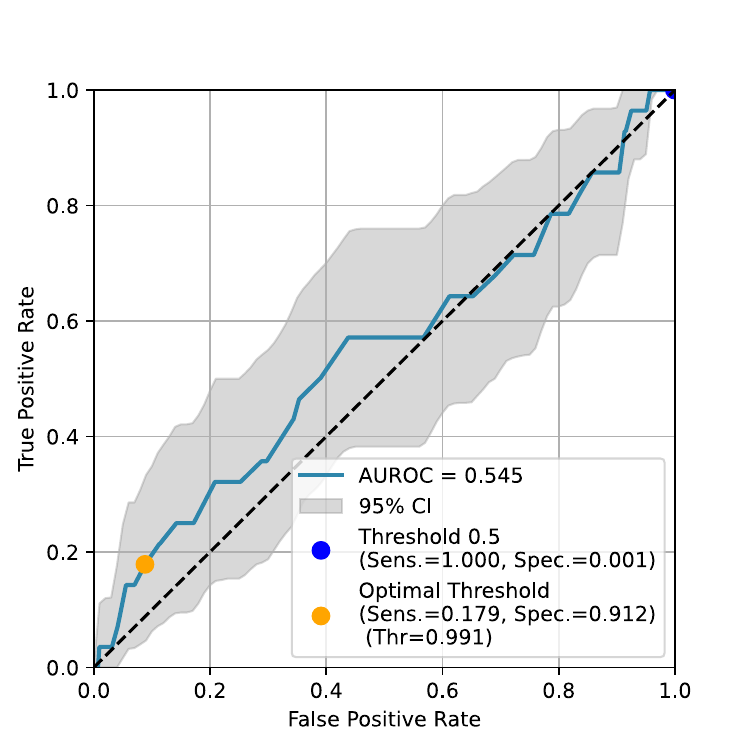}
        \caption{DERM12345 - Indeterminate Diagnosis}
        \label{fig:indeterminateROC_derm_27b}
    \end{subfigure}
    \hfill
    \begin{subfigure}[t]{0.3\textwidth}
        \centering
        \includegraphics[width=\linewidth]{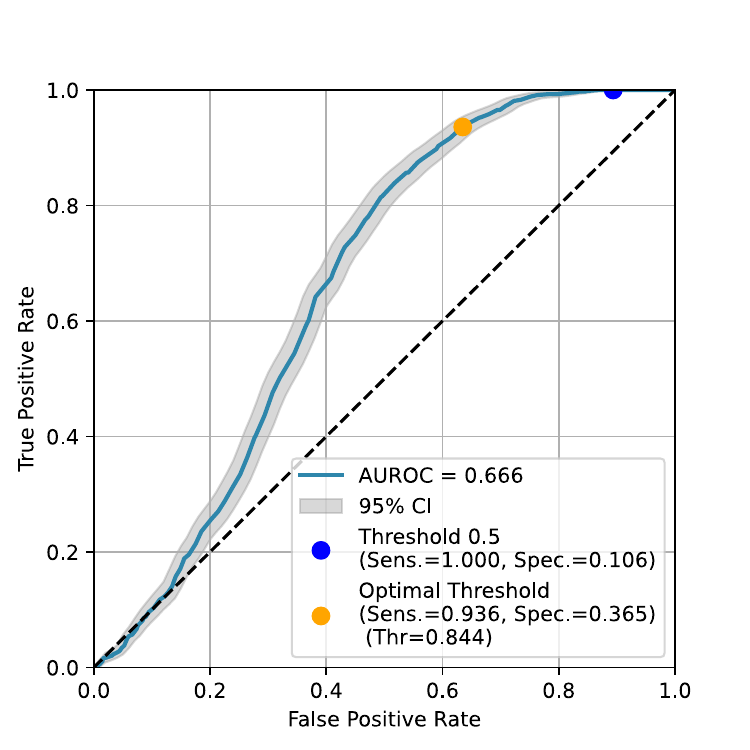}
        \caption{MIMIC-CXR - Support Devices}
        \label{fig:supportDevROC_mimic_27b}
    \end{subfigure}
    
    \caption{ROC curves for selected datasets and classes using MedGemma-27B. The default threshold of 0.5, corresponding to greedy sampling, is compared to the ``Optimal Threshold,'' which maximizes the sum of sensitivity and specificity, computed over the validation set.}
    \label{fig:roc_examples_27b}
\end{figure*}

Figure~\ref{fig:roc_examples_27b} illustrates how the choice of operating point determines the clinically relevant balance between sensitivity and specificity. It shows the ROC curves for several example classes, comparing the default threshold (0.5) with an ``optimal'' threshold that maximizes the sum of sensitivity and specificity (selected on the validation set). For the ``No Finding'' class on the CheXpert dataset (Fig.~\ref{fig:noFindingROC_chexpert_27b}), MedGemma-27B achieves a high AUROC of 0.921, indicating strong class separability. However, at the default threshold the model effectively labels all studies as abnormal (i.e., answers \no to ``No Finding''), yielding a sensitivity of 0.0 and a specificity of 1.0---an operating point that is clinically unhelpful because it fails to identify any radiographically normal examinations.
By shifting the threshold using the validation set to 0.005, the sensitivity increased to 0.945 while yielding a specificity of 0.696. 
Conversely, for the ``Support Devices'' class (Fig.~\ref{fig:supportDevROC_chexpert_27b}), the default threshold yielded a very high sensitivity of 0.994 but a poor specificity of 0.108.
Adjusting the threshold to 0.899 resulted in a sensitivity of 0.740 and a specificity of 0.819. For ``Pleural Effusion'' (Fig.~\ref{fig:pleuralEffusionROC_chexpert_27b}),  the implicit threshold is closer to the optimal point, but adjustment still offers improvement, moving from a sensitivity of 0.700 and specificity of 0.900 to a sensitivity of 0.850 and specificity of 0.792 at a threshold of 0.072. These examples demonstrated that the model's default operating point is often suboptimal and that the optimal threshold is highly class-dependent.

A comparison between the medically adapted MedGemma models and their general-purpose Gemma-3 counterparts underscored the value of domain-specific fine-tuning (Table \ref{tab:DifferentModels}). The performance gap was most pronounced for chest radiography; on CheXpert, for instance, the MedGemma-27B model achieved an average AUROC of 0.819, substantially outperforming the Gemma-3-27B's score of 0.535. Similar gains were observed on the MIMIC-CXR dataset. For histopathology classification on NCT-CRC-HE, the gain was pronounced for the 4B models (0.817 vs. 0.687) but small for the 27B models (0.783 vs. 0.769). Unexpectedly, however, the general-purpose models outperformed the MedGemma variants on the dermatology datasets (DERM12345 for both model sizes and HAM10000 for the 27B models), despite the latter having been fine-tuned on dermatological images~\cite{sellergren2025medgemmatechnicalreport}. Differences between models were not formally tested.
\subsection{Distribution of Optimal Thresholds}

\begin{figure}
    \centering
    \includegraphics[width=\linewidth]{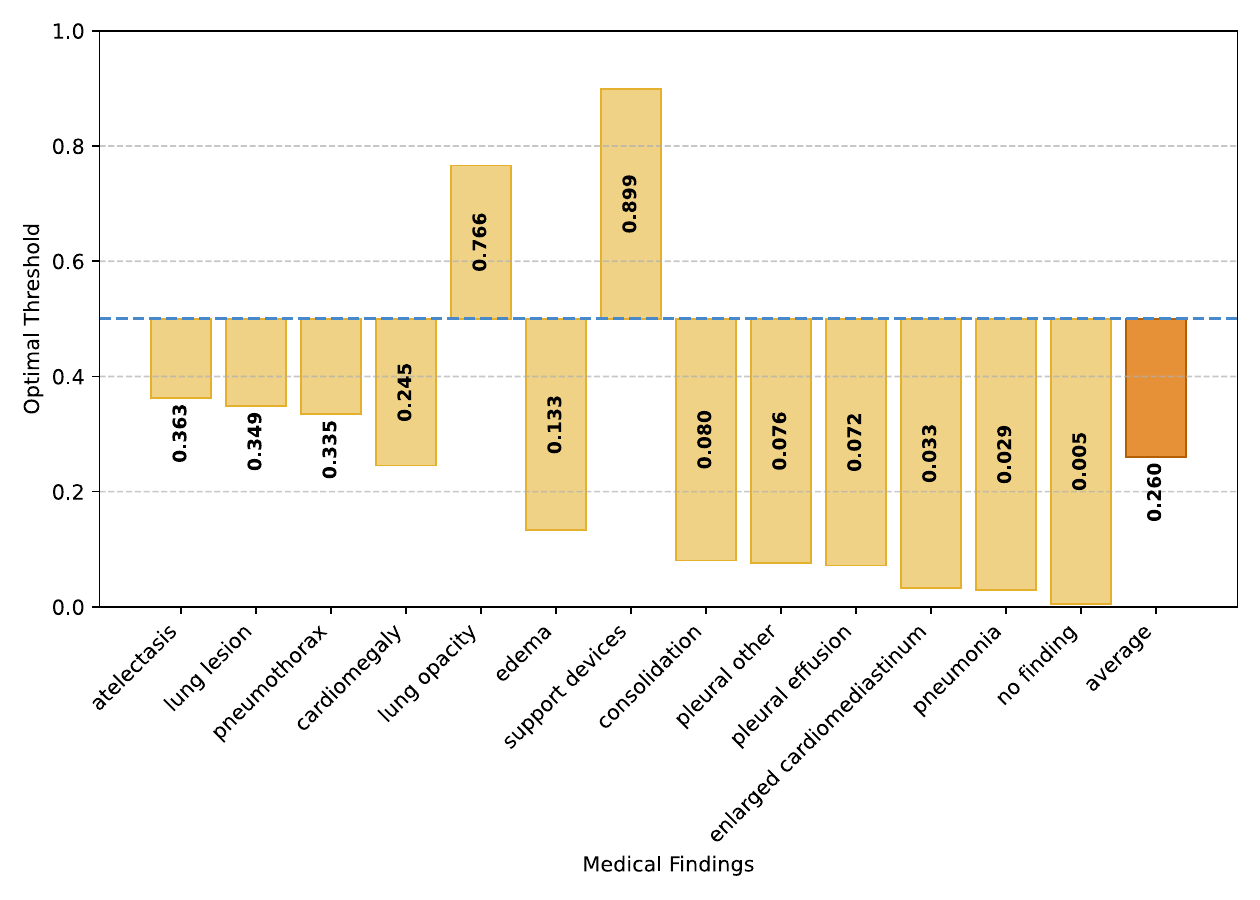}
    \caption{Distribution of optimal classification thresholds across CheXpert classes for MedGemma-27B. Optimal thresholds, defined as those maximizing sensitivity plus specificity, fall below 0.5 for 11 of 13 findings. Bars show the distance from the default 0.5 threshold, terminating at the optimal threshold value. The bar labeled ``average'' shows the mean across classes.}
    \label{fig:optThresholdDistribution}
\end{figure}

Figure~\ref{fig:optThresholdDistribution} illustrates the deviation of the optimal thresholds from the implicit default threshold of 0.5 across the different CheXpert classes.
For 11 of the 13 classes, the optimal threshold lies below the implicit threshold determined by the model's training.
This indicates that for these findings, the model exhibits a bias toward negative classification, resulting in higher specificity than sensitivity at the implicitly set threshold of 0.5.

\begin{table}[htbp]
\centering
\small
\caption{Average AUROC values with corresponding 95\% confidence intervals for different models using the normalized probabilities}
\label{tab:DifferentModels}
\begin{tabular}{lcccc}
\toprule
Dataset & MedGemma-27B & MedGemma-4B & Gemma-3-27B & Gemma-3-4B \\
\midrule
\multirow{2}{*}{CheXpert} & 0.819 & 0.790 & 0.535 & 0.519 \\
 & [0.794, 0.845] & [0.750, 0.827] & [0.496, 0.573] & [0.483, 0.557] \\
\addlinespace
\multirow{2}{*}{DERM12345} & 0.701 & 0.636 & 0.749 & 0.680 \\
 & [0.682, 0.719] & [0.610, 0.671] & [0.730, 0.777] & [0.660, 0.704] \\
\addlinespace
\multirow{2}{*}{G1020} & 0.494 & 0.541 & 0.507 & 0.480 \\
 & [0.437, 0.551] & [0.480, 0.603] & [0.446, 0.567] & [0.422, 0.538] \\
\addlinespace
\multirow{2}{*}{HAM10000} & 0.654 & 0.588 & 0.681 & 0.589 \\
 & [0.642, 0.666] & [0.572, 0.603] & [0.669, 0.692] & [0.575, 0.602] \\
\addlinespace
\multirow{2}{*}{MIMIC} & 0.658 & 0.673 & 0.533 & 0.514 \\
 & [0.647, 0.669] & [0.662, 0.684] & [0.522, 0.544] & [0.503, 0.525] \\
\addlinespace
\multirow{2}{*}{NCT-CRC-HE} & 0.783 & 0.817 & 0.769 & 0.687 \\
 & [0.777, 0.788] & [0.811, 0.822] & [0.762, 0.775] & [0.680, 0.694] \\
\bottomrule
\end{tabular}
\vspace{0.5ex}
{\footnotesize CI = 95\% confidence interval. All values are based on the normalized token probabilities.}
\end{table}

\subsection{Inference Latency}
A single decision required a median of 54~ms (MedGemma-4B), 53~ms (Gemma-3-4B), 78~ms (MedGemma-27B), and 79~ms (Gemma-3-27B).
When all findings of an image were queried in a single call, the latency per finding decreased to 12--14~ms; for example, all 14 CheXpert findings of one radiograph were assessed in about 0.2~s.

\section{Discussion}\label{sec:discussion}

Our results demonstrate that the implicit classification threshold used by large language models is often suboptimal for binary classification tasks in medicine. By extracting and normalizing the probabilities of \yes and \no tokens, we can define a continuous score that allows for threshold selection, improving the diagnostic utility of these models. This simple, post-hoc method reveals that while a model may have strong discriminative capabilities (as measured by AUROC), its default output corresponds to a fixed operating point that is often poorly suited to the clinical task.

The comparison of the three approaches in Table~\ref{tab:medgemma27b_auroc_values} shows that normalizing the \yes and \no token probabilities yields the highest AUROC for MedGemma-27B on four of six datasets, whereas raw probabilities were slightly but significantly better on MIMIC-CXR and NCT-CRC-HE.
This suggests that the raw logit values for these two tokens contain the most relevant signal for the binary task, and that the influence of other vocabulary tokens is a source of noise.
For MedGemma-27B, the raw-probability approach achieved higher AUROC values than text-elicited probability estimates on all six datasets (not formally tested), but it is still influenced by probability mass assigned to irrelevant vocabulary tokens. The poorer performance of prompting the model to output a probability as text, despite constrained decoding, suggests that these models are not reliably calibrated to express their internal uncertainty as an accurate numeric value.
The observation that, for MedGemma-27B, raw token probabilities outperform direct text inference (for the other models, the ordering varied across datasets), which contrasts with the findings of Gu et al.~\cite{Gu2024Probabilistic}, may have two potential explanations. First, our evaluation uses different models and focuses exclusively on image-based tasks, whereas their work included text. Second, our use of guided decoding ensures a valid numerical output for every sample. In contrast, the unconstrained generation used by Gu et al. might yield non-numerical responses (e.g., ``I don't know''), which would be excluded from their analysis and could introduce a selection bias.

The class- and dataset-dependent nature of the optimal threshold, as seen in Figure~\ref{fig:roc_examples_27b}, highlights the necessity of selecting the decision threshold for specific use cases.

Figure~\ref{fig:optThresholdDistribution} shows the deviation between the implicit and optimal thresholds across CheXpert classes. For 11 of 13 classes, the optimal threshold falls below 0.5, indicating a pronounced model bias toward negative predictions. This misaligned default threshold has important clinical ramifications, as different applications demand different operating points. For instance, screening tasks require high sensitivity to minimize missed diagnoses, necessitating a lower threshold than confirmatory tests, where high specificity is essential to avoid false positives. These results demonstrate that no universal threshold exists across all classes and datasets; optimal threshold selection must be adapted to the specific clinical context and task requirements.

Our approach shares a central idea with recently introduced decision models such as Jev~\cite{typesafe2026jev}: rather than generating text, the model is queried once and returns the probability of a predefined answer that downstream software can threshold.
Unlike Jev, which is a proprietary, text-only model trained with a dedicated objective for calibrated decisions, our approach requires no additional training, uses open-weight models, and applies to image inputs.
Notably, early independent evaluations of Jev found that no single temperature fitted all tasks and recommended selecting the decision threshold on a local selection set and re-checking it on held-out items~\cite{li2026jevjudge}.
Similarly, a benchmark of Jev on 37 datasets found that its binary probabilities ranked cases well but were poorly placed relative to a fixed threshold of 0.5~\cite{deusser2026jev}. On four of these datasets, thresholds tuned on training data improved performance both for Jev and for open LLMs whose yes/no probabilities were read out and renormalized as in our study~\cite{deusser2026jev}.
This mirrors our observation that optimal thresholds vary strongly across classes and datasets, suggesting that local threshold selection remains necessary regardless of how the probabilities are produced.
In our setting, a single decision took a median of 53--79~ms on one GPU, which is comparable to the latencies reported for such decision models (70--500~ms~\cite{typesafe2026jev}), although these figures are not directly comparable, as the latter include the overhead of a hosted API.

The performance of MedGemma varies significantly across tasks.
While it shows strong performance on some classes, such as ``No Finding'' on CheXpert or ``Lymphocytes'' on NCT-CRC-HE, its performance on other tasks is close to random, indicating the current limitations of generalist medical models.
The AUROC analysis is crucial here, as it distinguishes between a poorly placed default threshold (a high AUROC with a suboptimal default threshold) and a genuine inability of the model to perform the task (an AUROC close to 0.5).

Future work could explore extending this method to a chain-of-thought setting.
One could potentially extract the probabilities from the final \yes or \no token after a reasoning process.
However, the reasoning itself would likely bias the final token probabilities, potentially skewing them heavily and diminishing the utility of the thresholding approach. Investigating this interaction remains an open question.

Our method relies on access to the token probabilities (or logits), which may not be available for closed-source models accessed via APIs. While one could theoretically estimate these probabilities by sampling many single-token answers, this would likely be impractical for most applications. Furthermore, this study was limited to binary classification. Extending this approach to multi-class problems would require further methodological development, for instance, by performing one-vs-rest classification for each class or by normalizing probabilities across a larger set of verbalizer tokens. For NCT-CRC-HE-100K, no mapping of patches to their source slides is publicly available, so validation and test patches may originate from the same slides. This does not affect the AUROC, which is computed on the test set alone, but the metrics at the validation-optimized threshold may be slightly optimistic. Furthermore, classification performance is sensitive to prompt formulation and structure. The model's true discriminative ability for certain classes may therefore exceed our observed results if suboptimal prompt design limits class separability.

In this paper, we have presented a simple yet effective method to improve the binary classification performance of large language models by shifting their decision threshold. We have shown that the default threshold, implicitly defined by greedy token sampling, is often suboptimal for medical applications.
By normalizing the probabilities of \yes and \no answer tokens, we can create a tunable score that allows for the selection of an appropriate operating point on the ROC curve, balancing sensitivity and specificity for the clinical task at hand. This approach enhances the reliability of LLMs in medical diagnostics and provides a framework for evaluating and deploying them more safely and effectively.

\section{Methods}\label{sec1}
All experiments were conducted using Python (version 3.12.3) and the vLLM library (version 0.9.2) for model inference.
We evaluated four open-access models: the medically fine-tuned multimodal MedGemma 4B and MedGemma 27B~\cite{sellergren2025medgemmatechnicalreport}, alongside their architecture-matched general-purpose counterparts Gemma~3~4B and Gemma~3~27B~\cite{gemmateam2025gemma3technicalreport}.
All four models are based on the Gemma~3 architecture, with the MedGemma variants further adapted for medical data.

For each classification task, we formulated a prompt that queried the presence of a specific clinical finding (see supplementary material for details), instructing the model to answer with a single token: \yes or \no.
This setup led the model to assign a high probability to one of the two designated verbalizer tokens.
The model input $X$ consisted of the task-specific prompt concatenated with an image. %
Decoding was constrained to a single token and terminated immediately.
From the first-step distribution, we extracted the raw token probabilities 
$p_{\text{raw}}(\mathtt{yes}\!\mid\!X)$ and $p_{\text{raw}}(\mathtt{no}\!\mid\!X)$, 
which are the probabilities assigned by the model to the $\mathtt{yes}$ or $\mathtt{no}$ token 
given the input $X$.
To account for any residual probability mass on tokens outside this binary set~\cite{hegselmann23a}, we normalized the probabilities over the two relevant tokens:
\begin{equation}
p_{\text{norm}}(\mathtt{yes}\!\mid\!X)=\frac{p_{\text{raw}}(\mathtt{yes}\!\mid\!X)}{p_{\text{raw}}(\mathtt{yes}\!\mid\!X)+p_{\text{raw}}(\mathtt{no}\!\mid\!X)}.
\end{equation}
This is equivalent to a two-class softmax function applied only to the logits of \yes and \no.
The complementary probability, $p_{\text{norm}}(\mathtt{no}\!\mid\!X)$, is calculated analogously, and the two probabilities sum to one by construction.
Throughout our analysis, we treated $p_{\text{norm}}(\mathtt{yes}\!\mid\!X)$ as the predicted probability of the positive class.
Since the denominator is shared, $p_{\text{norm}}(\mathtt{yes}\!\mid\!X)>0.5$ if and only if $p_{\text{raw}}(\mathtt{yes}\!\mid\!X)>p_{\text{raw}}(\mathtt{no}\!\mid\!X)$.
Thus, greedy decoding, for example by setting the temperature to 0 or top-k to 1, is equivalent to applying a fixed classification threshold of 0.5.

We used $p_{\text{norm}}(\mathtt{yes}\!\mid\!X)$ as a continuous score to sweep the decision threshold across the interval $[0,1]$, which allowed us to generate ROC curves and compute the AUROC as a threshold-independent measure of performance~\cite{Bradley97AUROC}.
For a fixed threshold, we reported accuracy as \(\frac{TP+TN}{TP+TN+FP+FN}\), sensitivity (true positive rate) as \(\frac{TP}{TP+FN}\), and specificity (true negative rate) as \(\frac{TN}{TN+FP}\), where \(TP\) denotes true positives, 
\(TN\) true negatives, \(FP\) false positives, and \(FN\) false negatives.

In addition to this token-probability-based scoring, we evaluated a complementary method evaluated by Gu et al.~\cite{Gu2024Probabilistic}, which involves explicitly prompting the model to output a numerical probability for the presence of a finding as text.
To ensure machine-readability and prevent cases where no valid probability could be extracted, we employed a constrained JSON template (see Supplement) that permitted only numeric outputs with three decimal places (all observed values lay between 0 and 1).
We also included results using the raw probability, $p_{\text{raw}}(\mathtt{yes}\!\mid\!X)$, directly. To prevent bias from intra-context leakage, we elicited only a single probability within any given context window, as requesting multiple probabilities or classes in the same context could allow the model to condition later outputs on earlier ones.

\subsection{Datasets}\label{subsec:datasets}
For datasets with more than 10{,}000 images, we randomly subsampled 10{,}000 instances to keep the evaluation computationally tractable.
We then split this subsample into 5{,}000 validation instances (used only to select the decision threshold) and 5{,}000 test instances (used only for the final performance evaluation).
If a dataset provides official validation and test splits, we used those; otherwise, we used all available data and created random validation/test splits as described above.

For chest radiography, we evaluated our method on two datasets: CheXpert~\cite{irvin2019Chexpert} and MIMIC-CXR-JPG~\cite{johnson2019mimicCxrJpg,johnson2019mimic,goldberger2000physionet}.
From CheXpert, we included the official test set of 668 images, for which the reference standard was established by a majority vote of five board-certified radiologists, alongside 234 validation images.
As the label ``Fracture'' has no positive cases in the validation split, no decision threshold could be derived for it. We therefore included all 14 CheXpert labels in the AUROC analysis but excluded ``Fracture'' from the threshold analysis, resulting in 13 labels there.
From MIMIC-CXR-JPG v2.0.0, we used the official test split, which comprises 3,269 studies (5,159 images), while the validation set consists of 1,808 studies (2,991 images).
For studies with multiple images (e.g., frontal and lateral views), all images of the study were provided to the model together.

For histopathology, we used NCT-CRC-HE-100K, which provides 100,000 non-overlapping H\&E-stained patches (\(224\times 224\) pixels at \(0.5\,\mu\mathrm{m/px}\)) from nine tissue classes~\cite{kather2018NctCrcHe100k}.
We sampled 10,000 patches for our evaluation, which were randomly split into 5,000 test and 5,000 validation images.
In ophthalmology, we used the full G1020 dataset, consisting of 1,020 high-resolution fundus images with glaucoma ground truth and pixel-level optic disc/cup annotations~\cite{bajwa2020g1020}.
It was split into 204 test and 816 validation images.
For dermatology, we used the HAM10000 dataset in its ISIC Archive version, which provides 11,720 dermatoscopic images with diagnoses verified by histopathology, follow-up, or expert consensus, organized into three hierarchical diagnosis levels with 18 classes in total~\cite{tschandl2018ham10000}. We sampled 10,000 images, which were randomly split into 5,000 test and 5,000 validation images.
Additionally, we used DERM12345, which provides 12,345 dermatoscopic images; we used the first three levels of its hierarchical diagnosis labels, comprising 37 classes~\cite{yilmaz2024derm12345}. We sampled 10,000 images, which were also randomly split into 5,000 test and 5,000 validation images.
For each dataset, we created binary ``yes/no'' prompts asking for each class.
Of the datasets used, MIMIC was the only one included in the training data for the MedGemma models~\cite{sellergren2025medgemmatechnicalreport}.
Conversely, the authors of MedGemma stated that the CheXpert dataset was not used at any stage of model development, which makes it an ideal benchmark for external validation.

\subsection{Statistical Analysis}\label{subsec:stats}
We estimated 95\% confidence intervals (CIs) and two-sided \(p\)-values using nonparametric bootstrap resampling with 10{,}000 replicates.
To adjust for multiple comparisons across 12 tests, we applied a Bonferroni correction, resulting in a per-comparison significance threshold of \(\alpha = 0.0042\).

\subsection{Inference Latency}\label{subsec:latency}
To quantify the computational cost of our approach, we measured the end-to-end latency of a single decision, i.e., one forward pass that generates a single token and returns the log-probabilities of \yes and \no, using vLLM on a single NVIDIA H100 GPU.
We used the prompts of all six datasets together with randomly sampled images from five datasets, with a new image for every query to prevent reuse of cached image tokens.
We report the median latency over 51 single queries per model, as well as the latency per finding when all findings of one image were queried in a single call.

\subsection{Ethics}
This retrospective study used only publicly available, de-identified datasets. Ethical approval and informed consent were therefore not required for this secondary analysis; access to MIMIC-CXR-JPG was obtained through PhysioNet in accordance with its data use agreement.

\subsection{Use of Generative AI}
Generative AI tools were used for language editing of the manuscript; all scientific content, data, analyses, and conclusions were produced and verified by the authors.

\subsection{Code Availability}
The code is publicly available at \url{https://github.com/TruhnLab/llm-threshold-calibration}.

\backmatter

\bmhead{Acknowledgements}
This study was funded by the Deutsche Forschungsgemeinschaft (DFG; 417508432, 517243167, 515639690), the German Federal Ministry of Research, Technology and Space (Transform Liver -- 031L0312C, DECIPHER-M -- 01KD2420B), the European Regional Development Fund (ERDF; KIMONA -- GA EFRE-20801783) and the European Union (Horizon Europe, ODELIA -- GA 101057091, ERC Starting Grant SAGMA -- GA 101222556). \textit{Funded by the European Union. Views and opinions expressed are however those of the author(s) only and do not necessarily reflect those of the European Union or the European Research Council Executive Agency. Neither the European Union nor the granting authority can be held responsible for them.} The funders played no role in study design, data collection, analysis and interpretation of data, or the writing of this manuscript. The authors gratefully acknowledge the computing time provided to them at the NHR Center NHR4CES at RWTH Aachen University (project number p0021834). This is funded by the Federal Ministry of Research, Technology and Space, and the state governments participating on the basis of the resolutions of the GWK for national high performance computing at universities (www.nhr-verein.de/unsere-partner). The data used in this publication was managed using the research data management platform Coscine (http://doi.org/10.17616/R31NJNJZ) with storage space of the Research Data Storage (RDS) (DFG: INST222/1261-1) and DataStorage.nrw (DFG: INST222/1530-1) granted by the DFG and Ministry of Culture and Science of the State of North Rhine-Westphalia. We used generative AI tools for language editing and rephrasing; all scientific content, data, analyses, and conclusions were written and verified by the authors.

\bmhead{Author Contributions}
P.W. conceived the study together with D.T., developed the methodology, implemented
the software, performed the formal analysis, conducted the experiments, curated the
data, prepared the figures and tables, and wrote the original draft. D.T. contributed
to the conceptualization, provided resources, supervised the project, handled project
administration, and acquired funding. S.N. supervised the project, handled project
administration, and acquired funding. F.H., C.K., and J.N.K. contributed to the review
and editing of the manuscript. All authors critically reviewed and edited the
manuscript and approved the final version.

\bmhead{Competing Interests}
D.T. holds shares in StratifAI GmbH, Germany, and Synagen GmbH, Germany, and has received honoraria for lectures from AstraZeneca, Bayer, Debiopharm, GE HealthCare, Gilead, MSD, Orthomol, Philips, and Roche. J.N.K. holds shares in StratifAI, Synagen, Spira Labs, Tremont AI, and Saterra AI; is Co-PI on institutional research grants from GSK and AstraZeneca; and declares honoraria or consulting fees from AstraZeneca, Bayer, Bioptimus, Daiichi Sankyo, Eisai, Janssen, Merck, MSD, Novartis, BMS, Roche, and Pfizer. All other authors declare no financial or non-financial competing interests.

\bmhead{Data Availability}
This study used only publicly available datasets; no new data were generated.
CheXpert~\cite{irvin2019Chexpert}, MIMIC-CXR-JPG~\cite{johnson2019mimicCxrJpg,johnson2019mimic,goldberger2000physionet},
NCT-CRC-HE-100K~\cite{kather2018NctCrcHe100k}, G1020~\cite{bajwa2020g1020},
HAM10000~\cite{tschandl2018ham10000}, and DERM12345~\cite{yilmaz2024derm12345}
are available from their respective original sources as cited. MIMIC-CXR-JPG requires credentialed access via PhysioNet (\url{https://physionet.org/}). The code is publicly available at \url{https://github.com/TruhnLab/llm-threshold-calibration}.



%

\clearpage
\setcounter{section}{0}
\setcounter{table}{0}
\renewcommand{\thesection}{S\arabic{section}}
\renewcommand{\theHsection}{S\arabic{section}}
\renewcommand{\thetable}{S\arabic{table}}
\renewcommand{\theHtable}{S\arabic{table}}
\begin{center}
{\Large\bfseries Supplementary Information}
\end{center}
\section{Code}
\urlstyle{tt}
The implementation and further details can be found at:\\ \url{https://github.com/TruhnLab/llm-threshold-calibration}

\section{Prompts}\label{sec:prompts}
Each label was queried in a separate context. The placeholder \texttt{\{finding\}} was replaced by the label phrase listed in Section~\ref{sec:labels}. The image was passed together with the prompt.

\subsection{Yes/no prompts (normalized and raw token probabilities)}
\noindent\begin{minipage}{\linewidth}
\textbf{CheXpert, MIMIC-CXR}
\begin{lstlisting}
You are a medical image analysis assistant.

You will be provided with medical imaging and one clinical statement.  

TASK  
Decide whether the imaging evidence supports the clinical statement **exactly as written**.

Clinical statement: The imaging shows {finding}.

If the imaging evidence supports the statement, answer with: yes  
If the imaging evidence contradicts the statement or is indeterminate, answer with: no
\end{lstlisting}
\end{minipage}
\medskip

\noindent\begin{minipage}{\linewidth}
\textbf{HAM10000, DERM12345}
\begin{lstlisting}
You are a medical image analysis assistant.

TASK
Based on the visual evidence in the image, answer the following question with "yes" or "no".

Question: Does the lesion in the image show features of {finding}?
\end{lstlisting}
\end{minipage}
\medskip

\noindent\begin{minipage}{\linewidth}
\textbf{NCT-CRC-HE-100K}
\begin{lstlisting}
You are a digital pathology assistant.

TASK
Based on the visual evidence in the image, answer the following question with "yes" or "no".

Question: Is the dominant tissue type in this image patch {finding}?
\end{lstlisting}
\end{minipage}
\medskip

\noindent\begin{minipage}{\linewidth}
\textbf{G1020}
\begin{lstlisting}
You are a retinal image analysis assistant.

TASK
Based on the visual evidence in the image, answer the following question with "yes" or "no".

Question: Does the image show signs of {finding}?
\end{lstlisting}
\end{minipage}
\medskip

\subsection{Prompts for text-elicited probabilities}
\noindent\begin{minipage}{\linewidth}
\textbf{CheXpert, MIMIC-CXR, HAM10000, DERM12345}
\begin{lstlisting}
You are a medical image analysis assistant.

TASK
Based on the visual evidence in the image, estimate the probability that the following finding is present.
Provide the probability as a JSON object with a single key "probability" and a value between 0.000 and 1.000.

Finding: {finding}
\end{lstlisting}
\end{minipage}
\medskip

\noindent\begin{minipage}{\linewidth}
\textbf{NCT-CRC-HE-100K}
\begin{lstlisting}
You are a digital pathology assistant.

TASK
Based on the visual evidence in the image, estimate the probability that the following finding is present.
Provide the probability as a JSON object with a single key "probability" and a value between 0.000 and 1.000.

Finding: Is the dominant tissue type in this image patch {finding}?
\end{lstlisting}
\end{minipage}
\medskip

\noindent\begin{minipage}{\linewidth}
\textbf{G1020}
\begin{lstlisting}
You are a retinal image analysis assistant.

TASK
Based on the visual evidence in the image, estimate the probability that the following finding is present.
Provide the probability as a JSON object with a single key "probability" and a value between 0.000 and 1.000.

Finding: {finding}
\end{lstlisting}
\end{minipage}
\medskip

\subsection{JSON schema for guided decoding}
The text-elicited probability was generated with guided decoding (xgrammar) constrained to the following JSON schema, which only admits numbers with three decimal places (all observed values lay between 0 and 1):
\begin{lstlisting}
{"type": "object",
 "properties": {"probability": {"type": "string", "pattern": "[01]\\.\\d{3}"}},
 "required": ["probability"]}
\end{lstlisting}

\section{Labels}\label{sec:labels}
For each dataset, the original label name and the phrase inserted into the prompts are listed. For HAM10000 and DERM12345, the level refers to the hierarchical diagnosis label (1: benign, indeterminate or malignant; 2 and 3: increasingly specific diagnoses).

\begin{table}[htbp]
\centering
\small
\caption{Labels of CheXpert (14 labels).}\label{tab:labels_CheXpert}
\begin{tabular}{p{0.4\linewidth}p{0.45\linewidth}}
\toprule
Original label & Phrase in prompt \\
\midrule
No Finding & no finding \\
Enlarged Cardiomediastinum & enlarged cardiomediastinum \\
Cardiomegaly & cardiomegaly \\
Lung Opacity & lung opacity \\
Lung Lesion & lung lesion \\
Edema & edema \\
Consolidation & consolidation \\
Pneumonia & pneumonia \\
Atelectasis & atelectasis \\
Pneumothorax & pneumothorax \\
Pleural Effusion & pleural effusion \\
Pleural Other & pleural other \\
Fracture & fracture \\
Support Devices & support devices \\
\bottomrule
\end{tabular}
\end{table}

\begin{table}[htbp]
\centering
\small
\caption{Labels of MIMIC-CXR (14 labels).}\label{tab:labels_MIMIC}
\begin{tabular}{p{0.4\linewidth}p{0.45\linewidth}}
\toprule
Original label & Phrase in prompt \\
\midrule
Atelectasis & atelectasis \\
Cardiomegaly & cardiomegaly \\
Consolidation & consolidation \\
Edema & edema \\
Enlarged Cardiomediastinum & enlarged cardiomediastinum \\
Fracture & fracture \\
Lung Lesion & lung lesion \\
Lung Opacity & lung opacity \\
No Finding & no finding \\
Pleural Effusion & pleural effusion \\
Pleural Other & pleural other \\
Pneumonia & pneumonia \\
Pneumothorax & pneumothorax \\
Support Devices & support devices \\
\bottomrule
\end{tabular}
\end{table}

\begin{table}[htbp]
\centering
\small
\caption{Labels of HAM10000 (18 labels).}\label{tab:labels_HAM10000}
\begin{tabular}{cp{0.38\linewidth}p{0.44\linewidth}}
\toprule
Level & Original label & Phrase in prompt \\
\midrule
1 & Benign & a benign skin lesion \\
1 & Indeterminate & a skin lesion with an indeterminate diagnosis \\
1 & Malignant & a malignant skin lesion \\
2 & Benign epidermal proliferations & benign epidermal proliferation \\
2 & Benign melanocytic proliferations & benign melanocytic proliferation \\
2 & Benign soft tissue proliferations - Fibro-histiocytic & benign fibro-histiocytic proliferation \\
2 & Benign soft tissue proliferations - Vascular & benign vascular proliferation \\
2 & Indeterminate epidermal proliferations & indeterminate epidermal proliferation requiring further evaluation \\
2 & Malignant adnexal epithelial proliferations - Follicular & malignant follicular adnexal epithelial proliferation \\
2 & Malignant epidermal proliferations & malignant epidermal proliferation \\
2 & Malignant melanocytic proliferations (Melanoma) & malignant melanocytic proliferation (melanoma) \\
3 & Basal cell carcinoma & basal cell carcinoma \\
3 & Dermatofibroma & dermatofibroma \\
3 & Melanoma, NOS & melanoma, not otherwise specified \\
3 & Nevus & a nevus (melanocytic nevus) \\
3 & Pigmented benign keratosis & pigmented benign keratosis (seborrheic keratosis) \\
3 & Solar or actinic keratosis & solar or actinic keratosis \\
3 & Squamous cell carcinoma, NOS & squamous cell carcinoma, not otherwise specified \\
\bottomrule
\end{tabular}
\end{table}

\begin{table}[htbp]
\centering
\small
\caption{Labels of DERM12345 (37 labels).}\label{tab:labels_DERM12345}
\begin{tabular}{cp{0.38\linewidth}p{0.44\linewidth}}
\toprule
Level & Original label & Phrase in prompt \\
\midrule
1 & Benign & a benign skin lesion \\
1 & Indeterminate & a skin lesion with an indeterminate diagnosis \\
1 & Malignant & a malignant skin lesion \\
2 & Benign epidermal proliferations & benign epidermal proliferation \\
2 & Benign melanocytic proliferations & benign melanocytic proliferation \\
2 & Benign soft tissue proliferations - Fibro-histiocytic & benign fibro-histiocytic soft tissue proliferation \\
2 & Benign soft tissue proliferations - Vascular & benign vascular soft tissue proliferation \\
2 & Flat melanotic pigmentations - not melanocytic nevus & flat melanotic pigmentation (not a melanocytic nevus) \\
2 & Indeterminate epidermal proliferations & indeterminate epidermal proliferation \\
2 & Malignant adnexal epithelial proliferations - Apocrine or Eccrine & malignant apocrine or eccrine adnexal epithelial proliferation \\
2 & Malignant adnexal epithelial proliferations - Follicular & malignant follicular adnexal epithelial proliferation \\
2 & Malignant epidermal proliferations & malignant epidermal proliferation \\
2 & Malignant melanocytic proliferations (Melanoma) & malignant melanocytic proliferation (melanoma) \\
2 & Malignant soft tissue proliferations - Fibro-histiocytic & malignant fibro-histiocytic soft tissue proliferation \\
2 & Malignant soft tissue proliferations - Vascular & malignant vascular soft tissue proliferation \\
3 & Angiokeratoma & angiokeratoma \\
3 & Basal cell carcinoma & basal cell carcinoma \\
3 & Dermatofibroma & dermatofibroma \\
3 & Dermatofibrosarcoma protuberans & dermatofibrosarcoma protuberans \\
3 & Hemangioma & hemangioma \\
3 & Ink-spot lentigo & ink-spot lentigo \\
3 & Kaposi sarcoma & Kaposi sarcoma \\
3 & Lentigo simplex & lentigo simplex \\
3 & Lichen planus like keratosis & lichen planus like keratosis \\
3 & Lymphangioma & lymphangioma \\
3 & Melanoma Invasive & invasive melanoma \\
3 & Melanoma in situ & melanoma in situ \\
3 & Melanoma, NOS & melanoma, not otherwise specified \\
3 & Nevus & nevus \\
3 & Paget disease & Paget disease \\
3 & Pyogenic granuloma & pyogenic granuloma \\
3 & Seborrheic keratosis & seborrheic keratosis \\
3 & Solar lentigo & solar lentigo \\
3 & Solar or actinic keratosis & solar or actinic keratosis \\
3 & Squamous cell carcinoma in situ & squamous cell carcinoma in situ \\
3 & Squamous cell carcinoma, NOS & squamous cell carcinoma, not otherwise specified \\
3 & Vascular spider & vascular spider \\
\bottomrule
\end{tabular}
\end{table}

\begin{table}[htbp]
\centering
\small
\caption{Labels of NCT-CRC-HE-100K (9 labels).}\label{tab:labels_NCT_CRC_HE_100K_1000}
\begin{tabular}{p{0.4\linewidth}p{0.45\linewidth}}
\toprule
Original label & Phrase in prompt \\
\midrule
ADI & adipose tissue \\
BACK & background elements \\
DEB & debris \\
LYM & lymphocytes \\
MUC & mucus \\
MUS & smooth muscle \\
NORM & normal colon mucosa \\
STR & cancer-associated stroma \\
TUM & colorectal adenocarcinoma epithelial cells \\
\bottomrule
\end{tabular}
\end{table}

\begin{table}[htbp]
\centering
\small
\caption{Labels of G1020 (2 labels).}\label{tab:labels_G1020}
\begin{tabular}{p{0.4\linewidth}p{0.45\linewidth}}
\toprule
Original label & Phrase in prompt \\
\midrule
glaucoma & glaucoma \\
normal & a normal eye \\
\bottomrule
\end{tabular}
\end{table}

\clearpage
\section{Per-class AUROC values}\label{sec:perclass}
AUROC values with 95\% confidence intervals on the test set for MedGemma-27B, for the normalized token probability, the raw token probability and the text-elicited probability.

\begin{table}[htbp]
\centering
\footnotesize
\setlength{\tabcolsep}{4pt}
\caption{Per-class AUROC values for CheXpert (MedGemma-27B, test set).}\label{tab:auroc_CheXpert}
\begin{tabular}{p{0.3\linewidth}ccc}
\toprule
Label & Normalized (95\% CI) & Raw (95\% CI) & Text-elicited (95\% CI) \\
\midrule
No Finding & 0.921 [0.892, 0.945] & 0.909 [0.881, 0.934] & 0.882 [0.852, 0.909] \\
Enlarged Cardiomediastinum & 0.809 [0.773, 0.842] & 0.769 [0.730, 0.805] & 0.733 [0.700, 0.764] \\
Cardiomegaly & 0.848 [0.812, 0.882] & 0.820 [0.781, 0.858] & 0.824 [0.787, 0.858] \\
Lung Opacity & 0.808 [0.773, 0.840] & 0.659 [0.616, 0.699] & 0.731 [0.702, 0.761] \\
Lung Lesion & 0.863 [0.763, 0.947] & 0.824 [0.697, 0.934] & 0.695 [0.552, 0.821] \\
Edema & 0.839 [0.795, 0.880] & 0.839 [0.793, 0.881] & 0.830 [0.786, 0.872] \\
Consolidation & 0.808 [0.731, 0.879] & 0.766 [0.670, 0.855] & 0.781 [0.693, 0.867] \\
Pneumonia & 0.806 [0.672, 0.918] & 0.802 [0.652, 0.929] & 0.750 [0.599, 0.878] \\
Atelectasis & 0.795 [0.760, 0.829] & 0.766 [0.727, 0.803] & 0.743 [0.709, 0.775] \\
Pneumothorax & 0.923 [0.873, 0.965] & 0.920 [0.862, 0.965] & 0.723 [0.549, 0.897] \\
Pleural Effusion & 0.902 [0.874, 0.928] & 0.891 [0.856, 0.922] & 0.845 [0.804, 0.884] \\
Pleural Other & 0.606 [0.444, 0.782] & 0.541 [0.354, 0.740] & 0.478 [0.404, 0.627] \\
Fracture & 0.688 [0.475, 0.911] & 0.685 [0.408, 0.942] & 0.629 [0.451, 0.865] \\
Support Devices & 0.855 [0.826, 0.883] & 0.684 [0.643, 0.723] & 0.756 [0.725, 0.787] \\
\bottomrule
\end{tabular}
\end{table}

\begin{table}[htbp]
\centering
\footnotesize
\setlength{\tabcolsep}{4pt}
\caption{Per-class AUROC values for MIMIC-CXR (MedGemma-27B, test set).}\label{tab:auroc_MIMIC}
\begin{tabular}{p{0.3\linewidth}ccc}
\toprule
Label & Normalized (95\% CI) & Raw (95\% CI) & Text-elicited (95\% CI) \\
\midrule
Atelectasis & 0.511 [0.488, 0.533] & 0.535 [0.513, 0.557] & 0.557 [0.538, 0.576] \\
Cardiomegaly & 0.710 [0.691, 0.729] & 0.695 [0.676, 0.713] & 0.704 [0.687, 0.721] \\
Consolidation & 0.681 [0.646, 0.715] & 0.684 [0.649, 0.718] & 0.668 [0.632, 0.703] \\
Edema & 0.741 [0.721, 0.760] & 0.731 [0.712, 0.750] & 0.710 [0.692, 0.728] \\
Enlarged Cardiomediastinum & 0.434 [0.386, 0.482] & 0.474 [0.425, 0.524] & 0.459 [0.417, 0.502] \\
Fracture & 0.669 [0.614, 0.722] & 0.652 [0.598, 0.704] & 0.526 [0.489, 0.566] \\
Lung Lesion & 0.644 [0.594, 0.693] & 0.617 [0.568, 0.667] & 0.590 [0.543, 0.636] \\
Lung Opacity & 0.588 [0.567, 0.609] & 0.631 [0.611, 0.651] & 0.579 [0.564, 0.594] \\
No Finding & 0.712 [0.687, 0.737] & 0.718 [0.694, 0.741] & 0.682 [0.660, 0.704] \\
Pleural Effusion & 0.838 [0.824, 0.852] & 0.833 [0.819, 0.847] & 0.802 [0.786, 0.817] \\
Pleural Other & 0.646 [0.571, 0.722] & 0.627 [0.558, 0.698] & 0.621 [0.557, 0.685] \\
Pneumonia & 0.622 [0.591, 0.654] & 0.609 [0.579, 0.641] & 0.597 [0.569, 0.625] \\
Pneumothorax & 0.748 [0.703, 0.790] & 0.782 [0.740, 0.821] & 0.637 [0.591, 0.685] \\
Support Devices & 0.666 [0.647, 0.684] & 0.732 [0.715, 0.749] & 0.618 [0.600, 0.636] \\
\bottomrule
\end{tabular}
\end{table}

\begin{table}[htbp]
\centering
\footnotesize
\setlength{\tabcolsep}{4pt}
\caption{Per-class AUROC values for HAM10000 (MedGemma-27B, test set).}\label{tab:auroc_HAM10000}
\begin{tabular}{p{0.3\linewidth}ccc}
\toprule
Label & Normalized (95\% CI) & Raw (95\% CI) & Text-elicited (95\% CI) \\
\midrule
Benign & 0.610 [0.590, 0.629] & 0.586 [0.567, 0.606] & 0.636 [0.618, 0.653] \\
Indeterminate & 0.427 [0.359, 0.495] & 0.555 [0.472, 0.636] & 0.590 [0.534, 0.643] \\
Malignant & 0.573 [0.551, 0.596] & 0.546 [0.524, 0.569] & 0.524 [0.504, 0.544] \\
Benign epidermal proliferations & 0.664 [0.641, 0.687] & 0.585 [0.559, 0.611] & 0.435 [0.415, 0.455] \\
Benign melanocytic proliferations & 0.513 [0.496, 0.531] & 0.480 [0.463, 0.498] & 0.522 [0.516, 0.529] \\
Benign soft tissue proliferations - Fibro-histiocytic & 0.624 [0.571, 0.677] & 0.603 [0.545, 0.660] & 0.434 [0.375, 0.492] \\
Benign soft tissue proliferations - Vascular & 0.611 [0.553, 0.668] & 0.604 [0.537, 0.670] & 0.736 [0.692, 0.774] \\
Indeterminate epidermal proliferations & 0.814 [0.769, 0.856] & 0.600 [0.527, 0.669] & 0.444 [0.388, 0.501] \\
Malignant adnexal epithelial proliferations - Follicular & 0.738 [0.711, 0.765] & 0.694 [0.664, 0.725] & 0.651 [0.621, 0.681] \\
Malignant epidermal proliferations & 0.522 [0.466, 0.578] & 0.428 [0.374, 0.482] & 0.327 [0.280, 0.375] \\
Malignant melanocytic proliferations (Melanoma) & 0.667 [0.641, 0.693] & 0.628 [0.600, 0.655] & 0.575 [0.552, 0.599] \\
Basal cell carcinoma & 0.805 [0.781, 0.828] & 0.766 [0.740, 0.791] & 0.606 [0.579, 0.633] \\
Dermatofibroma & 0.588 [0.522, 0.655] & 0.621 [0.552, 0.688] & 0.526 [0.463, 0.590] \\
Melanoma, NOS & 0.627 [0.601, 0.654] & 0.602 [0.575, 0.630] & 0.560 [0.536, 0.585] \\
Nevus & 0.687 [0.671, 0.702] & 0.604 [0.588, 0.621] & 0.597 [0.586, 0.609] \\
Pigmented benign keratosis & 0.698 [0.676, 0.719] & 0.572 [0.548, 0.597] & 0.475 [0.460, 0.489] \\
Solar or actinic keratosis & 0.865 [0.834, 0.896] & 0.778 [0.730, 0.824] & 0.654 [0.617, 0.687] \\
Squamous cell carcinoma, NOS & 0.735 [0.694, 0.775] & 0.669 [0.622, 0.716] & 0.471 [0.427, 0.518] \\
\bottomrule
\end{tabular}
\end{table}

\begin{table}[htbp]
\centering
\footnotesize
\setlength{\tabcolsep}{4pt}
\caption{Per-class AUROC values for DERM12345 (MedGemma-27B, test set).}\label{tab:auroc_DERM12345}
\begin{tabular}{p{0.3\linewidth}ccc}
\toprule
Label & Normalized (95\% CI) & Raw (95\% CI) & Text-elicited (95\% CI) \\
\midrule
Benign & 0.835 [0.819, 0.849] & 0.793 [0.775, 0.810] & 0.839 [0.817, 0.859] \\
Indeterminate & 0.545 [0.425, 0.665] & 0.643 [0.541, 0.737] & 0.709 [0.614, 0.800] \\
Malignant & 0.492 [0.461, 0.523] & 0.476 [0.446, 0.506] & 0.500 [0.469, 0.531] \\
Benign epidermal proliferations & 0.642 [0.606, 0.677] & 0.604 [0.571, 0.638] & 0.488 [0.460, 0.516] \\
Benign melanocytic proliferations & 0.744 [0.724, 0.763] & 0.637 [0.616, 0.658] & 0.585 [0.571, 0.599] \\
Benign soft tissue proliferations - Fibro-histiocytic & 0.583 [0.529, 0.638] & 0.512 [0.455, 0.567] & 0.393 [0.340, 0.448] \\
Benign soft tissue proliferations - Vascular & 0.624 [0.571, 0.673] & 0.642 [0.589, 0.693] & 0.544 [0.505, 0.580] \\
Flat melanotic pigmentations - not melanocytic nevus & 0.949 [0.914, 0.981] & 0.807 [0.768, 0.846] & 0.648 [0.522, 0.770] \\
Indeterminate epidermal proliferations & 0.876 [0.828, 0.916] & 0.622 [0.531, 0.707] & 0.562 [0.499, 0.620] \\
Malignant adnexal epithelial proliferations - Apocrine or Eccrine & 0.758 [0.653, 0.838] & 0.594 [0.422, 0.740] & 0.886 [0.878, 0.894] \\
Malignant adnexal epithelial proliferations - Follicular & 0.832 [0.807, 0.856] & 0.759 [0.727, 0.790] & 0.749 [0.711, 0.786] \\
Malignant epidermal proliferations & 0.496 [0.445, 0.549] & 0.421 [0.368, 0.477] & 0.486 [0.442, 0.531] \\
Malignant melanocytic proliferations (Melanoma) & 0.575 [0.524, 0.627] & 0.480 [0.432, 0.529] & 0.407 [0.364, 0.451] \\
Malignant soft tissue proliferations - Fibro-histiocytic & 0.605 [0.525, 0.644] & 0.313 [0.140, 0.605] & 0.311 [0.235, 0.517] \\
Malignant soft tissue proliferations - Vascular & 0.792 [0.684, 0.890] & 0.842 [0.788, 0.904] & 0.668 [0.485, 0.817] \\
Angiokeratoma & 0.823 [0.748, 0.898] & 0.864 [0.776, 0.934] & 0.575 [0.491, 0.701] \\
Basal cell carcinoma & 0.874 [0.851, 0.896] & 0.844 [0.821, 0.865] & 0.766 [0.735, 0.796] \\
Dermatofibroma & 0.687 [0.629, 0.743] & 0.661 [0.609, 0.710] & 0.511 [0.455, 0.568] \\
Dermatofibrosarcoma protuberans & 0.753 [0.591, 0.828] & 0.301 [0.006, 0.535] & 0.167 [0.160, 0.173] \\
Hemangioma & 0.783 [0.724, 0.837] & 0.795 [0.741, 0.845] & 0.817 [0.773, 0.857] \\
Ink-spot lentigo & 0.305 [0.058, 0.556] & 0.362 [0.078, 0.649] & 0.244 [0.237, 0.251] \\
Kaposi sarcoma & 0.372 [0.227, 0.515] & 0.263 [0.132, 0.484] & 0.470 [0.466, 0.473] \\
Lentigo simplex & 0.766 [0.635, 0.880] & 0.725 [0.611, 0.817] & 0.515 [0.375, 0.658] \\
Lichen planus like keratosis & 0.672 [0.519, 0.848] & 0.615 [0.315, 0.973] & 0.526 [0.287, 0.862] \\
Lymphangioma & 0.842 [0.790, 0.888] & 0.915 [0.888, 0.940] & 0.500 [0.500, 0.500] \\
Melanoma Invasive & 0.568 [0.463, 0.670] & 0.542 [0.432, 0.648] & 0.319 [0.232, 0.412] \\
Melanoma in situ & 0.608 [0.504, 0.712] & 0.505 [0.394, 0.621] & 0.351 [0.270, 0.440] \\
Melanoma, NOS & 0.491 [0.419, 0.565] & 0.408 [0.344, 0.477] & 0.435 [0.368, 0.502] \\
Nevus & 0.835 [0.819, 0.850] & 0.770 [0.752, 0.788] & 0.676 [0.663, 0.690] \\
Paget disease & 0.311 [0.214, 0.414] & 0.230 [0.168, 0.287] & 0.279 [0.272, 0.285] \\
Pyogenic granuloma & 0.931 [0.902, 0.958] & 0.968 [0.947, 0.988] & 0.490 [0.488, 0.492] \\
Seborrheic keratosis & 0.735 [0.703, 0.766] & 0.744 [0.715, 0.773] & 0.575 [0.561, 0.588] \\
Solar lentigo & 0.820 [0.726, 0.897] & 0.714 [0.630, 0.795] & 0.603 [0.540, 0.646] \\
Solar or actinic keratosis & 0.890 [0.843, 0.933] & 0.837 [0.784, 0.888] & 0.751 [0.693, 0.805] \\
Squamous cell carcinoma in situ & 0.785 [0.705, 0.858] & 0.650 [0.539, 0.758] & 0.372 [0.280, 0.473] \\
Squamous cell carcinoma, NOS & 0.888 [0.866, 0.908] & 0.870 [0.840, 0.899] & 0.758 [0.718, 0.795] \\
Vascular spider & 0.872 [0.561, 0.999] & 0.863 [0.545, 0.991] & 0.723 [0.467, 0.989] \\
\bottomrule
\end{tabular}
\end{table}

\begin{table}[htbp]
\centering
\footnotesize
\setlength{\tabcolsep}{4pt}
\caption{Per-class AUROC values for NCT-CRC-HE-100K (MedGemma-27B, test set).}\label{tab:auroc_NCT_CRC_HE_100K_1000}
\begin{tabular}{p{0.3\linewidth}ccc}
\toprule
Label & Normalized (95\% CI) & Raw (95\% CI) & Text-elicited (95\% CI) \\
\midrule
ADI & 0.968 [0.960, 0.976] & 0.974 [0.967, 0.980] & 0.831 [0.821, 0.842] \\
BACK & 0.932 [0.922, 0.942] & 0.909 [0.897, 0.921] & 0.497 [0.495, 0.498] \\
DEB & 0.769 [0.746, 0.791] & 0.707 [0.683, 0.729] & 0.673 [0.651, 0.694] \\
LYM & 0.956 [0.950, 0.961] & 0.929 [0.922, 0.935] & 0.882 [0.876, 0.888] \\
MUC & 0.739 [0.717, 0.761] & 0.784 [0.763, 0.805] & 0.502 [0.500, 0.506] \\
MUS & 0.543 [0.521, 0.565] & 0.641 [0.619, 0.661] & 0.578 [0.561, 0.594] \\
NORM & 0.803 [0.783, 0.822] & 0.782 [0.763, 0.800] & 0.698 [0.675, 0.721] \\
STR & 0.536 [0.512, 0.561] & 0.659 [0.637, 0.681] & 0.491 [0.475, 0.508] \\
TUM & 0.796 [0.780, 0.812] & 0.779 [0.762, 0.795] & 0.629 [0.621, 0.637] \\
\bottomrule
\end{tabular}
\end{table}

\begin{table}[htbp]
\centering
\footnotesize
\setlength{\tabcolsep}{4pt}
\caption{Per-class AUROC values for G1020 (MedGemma-27B, test set).}\label{tab:auroc_G1020}
\begin{tabular}{p{0.3\linewidth}ccc}
\toprule
Label & Normalized (95\% CI) & Raw (95\% CI) & Text-elicited (95\% CI) \\
\midrule
glaucoma & 0.522 [0.438, 0.603] & 0.506 [0.419, 0.591] & 0.500 [0.500, 0.500] \\
normal & 0.467 [0.383, 0.549] & 0.471 [0.388, 0.556] & 0.473 [0.410, 0.538] \\
\bottomrule
\end{tabular}
\end{table}


\begin{thebibliography}{10}
\expandafter\ifx\csname url\endcsname\relax
  \def\url#1{\burl{#1}}\fi
\expandafter\ifx\csname urlprefix\endcsname\relax\def\urlprefix{URL }\fi
\providecommand{\bibinfo}[2]{#2}
\providecommand{\eprint}[2][]{\url{#2}}
\providecommand{\doi}[1]{\url{https://doi.org/#1}}
\bibcommenthead

\bibitem{zhou2024preX22}
\bibinfo{author}{Zhou, J.} \emph{et~al.}
\newblock \bibinfo{title}{Pre-trained multimodal large language model enhances
  dermatological diagnosis using skingpt-4}.
\newblock \emph{\bibinfo{journal}{Nature Communications}}
  \textbf{\bibinfo{volume}{15}}, \bibinfo{pages}{5649} (\bibinfo{year}{2024}).

\bibitem{Wang2024Calibrated}
\bibinfo{author}{Wang, C.}, \bibinfo{author}{Szarvas, G.},
  \bibinfo{author}{Balazs, G.}, \bibinfo{author}{Danchenko, P.} \&
  \bibinfo{author}{Ernst, P.}
\newblock \bibinfo{title}{Calibrating verbalized probabilities for large
  language models}.
\newblock \emph{\bibinfo{journal}{CoRR}}
  \textbf{\bibinfo{volume}{abs/2410.06707}} (\bibinfo{year}{2024}).
\newblock \urlprefix\url{https://doi.org/10.48550/arXiv.2410.06707}.

\bibitem{schick2021TextClassificationX1}
\bibinfo{author}{Schick, T.} \& \bibinfo{author}{Sch{\"{u}}tze, H.}
\newblock \bibinfo{editor}{Merlo, P.}, \bibinfo{editor}{Tiedemann, J.} \&
  \bibinfo{editor}{Tsarfaty, R.} (eds) \emph{\bibinfo{title}{Exploiting
  cloze-questions for few-shot text classification and natural language
  inference}}.
\newblock (eds \bibinfo{editor}{Merlo, P.}, \bibinfo{editor}{Tiedemann, J.} \&
  \bibinfo{editor}{Tsarfaty, R.}) \emph{\bibinfo{booktitle}{Proceedings of the
  16th Conference of the European Chapter of the Association for Computational
  Linguistics: Main Volume, {EACL} 2021, Online, April 19 - 23, 2021}},
  \bibinfo{pages}{255--269} (\bibinfo{publisher}{Association for Computational
  Linguistics}, \bibinfo{year}{2021}).
\newblock \urlprefix\url{https://doi.org/10.18653/v1/2021.eacl-main.20}.

\bibitem{hegselmann23a}
\bibinfo{author}{Hegselmann, S.} \emph{et~al.}
\newblock \bibinfo{editor}{Ruiz, F. J.~R.}, \bibinfo{editor}{Dy, J.~G.} \&
  \bibinfo{editor}{van~de Meent, J.} (eds) \emph{\bibinfo{title}{Tabllm:
  Few-shot classification of tabular data with large language models}}.
\newblock (eds \bibinfo{editor}{Ruiz, F. J.~R.}, \bibinfo{editor}{Dy, J.~G.} \&
  \bibinfo{editor}{van~de Meent, J.}) \emph{\bibinfo{booktitle}{International
  Conference on Artificial Intelligence and Statistics, 25-27 April 2023, Palau
  de Congressos, Valencia, Spain}}, Vol. \bibinfo{volume}{206} of
  \emph{\bibinfo{series}{Proceedings of Machine Learning Research}},
  \bibinfo{pages}{5549--5581} (\bibinfo{publisher}{{PMLR}},
  \bibinfo{year}{2023}).
\newblock
  \urlprefix\url{https://proceedings.mlr.press/v206/hegselmann23a.html}.

\bibitem{Giovannotti2024Calibrated}
\bibinfo{author}{Giovannotti, P.} \& \bibinfo{author}{Gammerman, A.}
\newblock \bibinfo{editor}{Vantini, S.}, \bibinfo{editor}{Fontana, M.},
  \bibinfo{editor}{Solari, A.}, \bibinfo{editor}{Bostr{\"{o}}m, H.} \&
  \bibinfo{editor}{Carlsson, L.} (eds) \emph{\bibinfo{title}{Calibrated large
  language models for binary question answering}}.
\newblock (eds \bibinfo{editor}{Vantini, S.}, \bibinfo{editor}{Fontana, M.},
  \bibinfo{editor}{Solari, A.}, \bibinfo{editor}{Bostr{\"{o}}m, H.} \&
  \bibinfo{editor}{Carlsson, L.}) \emph{\bibinfo{booktitle}{The 13th Symposium
  on Conformal and Probabilistic Prediction with Applications, 9-11 September
  2024, Politecnico di Milano, Milano, Italy}}, Vol. \bibinfo{volume}{230} of
  \emph{\bibinfo{series}{Proceedings of Machine Learning Research}},
  \bibinfo{pages}{218--235} (\bibinfo{publisher}{{PMLR}},
  \bibinfo{year}{2024}).
\newblock
  \urlprefix\url{https://proceedings.mlr.press/v230/giovannotti24a.html}.

\bibitem{cherian2024conformalPredictionX12}
\bibinfo{author}{Cherian, J.~J.}, \bibinfo{author}{Gibbs, I.} \&
  \bibinfo{author}{Cand{\`{e}}s, E.~J.}
\newblock \bibinfo{editor}{Globersons, A.} \emph{et~al.} (eds)
  \emph{\bibinfo{title}{Large language model validity via enhanced conformal
  prediction methods}}.
\newblock (eds \bibinfo{editor}{Globersons, A.} \emph{et~al.})
  \emph{\bibinfo{booktitle}{Advances in Neural Information Processing Systems
  38: Annual Conference on Neural Information Processing Systems 2024, NeurIPS
  2024, Vancouver, BC, Canada, December 10 - 15, 2024}} (\bibinfo{year}{2024}).
\newblock
  \urlprefix\url{http://papers.nips.cc/paper\_files/paper/2024/hash/d02ff1aeaa5c268dc34790dd1ad21526-Abstract-Conference.html}.

\bibitem{zhao2021CalibrationX6}
\bibinfo{author}{Zhao, Z.}, \bibinfo{author}{Wallace, E.},
  \bibinfo{author}{Feng, S.}, \bibinfo{author}{Klein, D.} \&
  \bibinfo{author}{Singh, S.}
\newblock \bibinfo{editor}{Meila, M.} \& \bibinfo{editor}{Zhang, T.} (eds)
  \emph{\bibinfo{title}{Calibrate before use: Improving few-shot performance of
  language models}}.
\newblock (eds \bibinfo{editor}{Meila, M.} \& \bibinfo{editor}{Zhang, T.})
  \emph{\bibinfo{booktitle}{Proceedings of the 38th International Conference on
  Machine Learning, {ICML} 2021, 18-24 July 2021, Virtual Event}}, Vol.
  \bibinfo{volume}{139} of \emph{\bibinfo{series}{Proceedings of Machine
  Learning Research}}, \bibinfo{pages}{12697--12706}
  (\bibinfo{publisher}{{PMLR}}, \bibinfo{year}{2021}).
\newblock \urlprefix\url{http://proceedings.mlr.press/v139/zhao21c.html}.

\bibitem{zhou2024BatchCalibrationX7}
\bibinfo{author}{Zhou, H.} \emph{et~al.} \emph{\bibinfo{title}{Batch
  calibration: Rethinking calibration for in-context learning and prompt
  engineering}}.
\newblock \emph{\bibinfo{booktitle}{The Twelfth International Conference on
  Learning Representations, {ICLR} 2024, Vienna, Austria, May 7-11, 2024}}
  (\bibinfo{publisher}{OpenReview.net}, \bibinfo{year}{2024}).
\newblock \urlprefix\url{https://openreview.net/forum?id=L3FHMoKZcS}.

\bibitem{dong2024xgrammar}
\bibinfo{author}{Dong, Y.} \emph{et~al.}
\newblock \bibinfo{title}{Xgrammar: Flexible and efficient structured
  generation engine for large language models}.
\newblock \emph{\bibinfo{journal}{CoRR}}
  \textbf{\bibinfo{volume}{abs/2411.15100}} (\bibinfo{year}{2024}).
\newblock \urlprefix\url{https://doi.org/10.48550/arXiv.2411.15100}.

\bibitem{typesafe2026jev}
\bibinfo{author}{Almeida, D.}
\newblock \bibinfo{title}{Introducing {System One} models and {Jev}}.
\newblock \bibinfo{howpublished}{TypeSafe AI Blog} (\bibinfo{year}{2026}).
\newblock
  \urlprefix\url{https://typesafe.ai/blog/introducing-system-one-models-and-jev}.
\newblock \bibinfo{note}{Accessed 29 September 2026}.

\bibitem{kahneman2011thinking}
\bibinfo{author}{Kahneman, D.}
\newblock \emph{\bibinfo{title}{Thinking, Fast and Slow}}
  (\bibinfo{publisher}{Farrar, Straus and Giroux}, \bibinfo{address}{New York},
  \bibinfo{year}{2011}).

\bibitem{madrid2026jevmedicine}
\bibinfo{author}{Madrid-Garc{\'i}a, A.} \& \bibinfo{author}{Merino-Barbancho,
  B.}
\newblock \bibinfo{title}{{Jev} in medicine: A benchmark evaluation}
  (\bibinfo{year}{2026}).
\newblock
  \bibinfo{eprint}{{\href{https://arxiv.org/abs/2609.34024}{{arXiv:2609.34024}}}}.

\bibitem{du2026systemonemedicine}
\bibinfo{author}{Du, Y.} \emph{et~al.}
\newblock \bibinfo{title}{How reliable are {System-One} decisions in medicine?
  {A} comparative study of {Jev}, {Jev}-like models, and {LLMs}}
  (\bibinfo{year}{2026}).
\newblock
  \bibinfo{eprint}{{\href{https://arxiv.org/abs/2610.18529}{{arXiv:2610.18529}}}}.

\bibitem{tang2026omnimedjev}
\bibinfo{author}{Tang, L.} \& \bibinfo{author}{Chen, C.}
\newblock \bibinfo{title}{{OmniMed-Jev}: Calibrating {LVLM} confidence for
  trustworthy medical multimodal decisions via {System One}}
  (\bibinfo{year}{2026}).
\newblock
  \bibinfo{eprint}{{\href{https://arxiv.org/abs/2610.00381}{{arXiv:2610.00381}}}}.

\bibitem{tang2026typedaudit}
\bibinfo{author}{Tang, L.} \& \bibinfo{author}{Zheng, Y.}
\newblock \bibinfo{title}{Typed decision models: An early evidence audit and
  evaluation checklist} (\bibinfo{year}{2026}).
\newblock
  \bibinfo{eprint}{{\href{https://arxiv.org/abs/2609.32160}{{arXiv:2609.32160}}}}.

\bibitem{li2026llmasjev}
\bibinfo{author}{Li, Y.} \& \bibinfo{author}{Wagle, J.}
\newblock \bibinfo{title}{{LLM}-as-{Jev}: {LLMs} are already {Jev}-style
  decision models -- when and how to fine-tune them} (\bibinfo{year}{2026}).
\newblock
  \bibinfo{eprint}{{\href{https://arxiv.org/abs/2610.02076}{{arXiv:2610.02076}}}}.

\bibitem{vickers2016netX25}
\bibinfo{author}{Vickers, A.~J.}, \bibinfo{author}{Van~Calster, B.} \&
  \bibinfo{author}{Steyerberg, E.~W.}
\newblock \bibinfo{title}{Net benefit approaches to the evaluation of
  prediction models, molecular markers, and diagnostic tests}.
\newblock \emph{\bibinfo{journal}{bmj}} \textbf{\bibinfo{volume}{352}}
  (\bibinfo{year}{2016}).

\bibitem{sellergren2025medgemmatechnicalreport}
\bibinfo{author}{Sellergren, A.} \emph{et~al.}
\newblock \bibinfo{title}{{MedGemma} technical report} (\bibinfo{year}{2025}).
\newblock \urlprefix\url{https://arxiv.org/abs/2507.05201}.
\newblock
  \bibinfo{eprint}{{\href{https://arxiv.org/abs/2507.05201}{{arXiv:2507.05201}}}}.

\bibitem{gemmateam2025gemma3technicalreport}
\bibinfo{author}{Team, G.} \emph{et~al.}
\newblock \bibinfo{title}{Gemma 3 technical report} (\bibinfo{year}{2025}).
\newblock \urlprefix\url{https://arxiv.org/abs/2503.19786}.
\newblock
  \bibinfo{eprint}{{\href{https://arxiv.org/abs/2503.19786}{{arXiv:2503.19786}}}}.

\bibitem{Gu2024Probabilistic}
\bibinfo{author}{Gu, B.}, \bibinfo{author}{Desai, R.~J.}, \bibinfo{author}{Lin,
  K.~J.} \& \bibinfo{author}{Yang, J.}
\newblock \bibinfo{title}{Probabilistic medical predictions of large language
  models}.
\newblock \emph{\bibinfo{journal}{npj Digit. Medicine}}
  \textbf{\bibinfo{volume}{7}} (\bibinfo{year}{2024}).
\newblock \urlprefix\url{https://doi.org/10.1038/s41746-024-01366-4}.

\bibitem{li2026jevjudge}
\bibinfo{author}{Li, Y.}, \bibinfo{author}{Miao, Y.},
  \bibinfo{author}{Krishnan, R.} \& \bibinfo{author}{Padman, R.}
\newblock \bibinfo{title}{{JEV}-as-a-judge: Accept when confident, escalate
  when unsure} (\bibinfo{year}{2026}).
\newblock
  \bibinfo{eprint}{{\href{https://arxiv.org/abs/2609.26550}{{arXiv:2609.26550}}}}.

\bibitem{deusser2026jev}
\bibinfo{author}{Deu{\ss}er, T.}, \bibinfo{author}{Sparrenberg, L.} \&
  \bibinfo{author}{Sifa, R.}
\newblock \bibinfo{title}{Evaluating and benchmarking the {System One} model
  {Jev}} (\bibinfo{year}{2026}).
\newblock
  \bibinfo{eprint}{{\href{https://arxiv.org/abs/2609.37647}{{arXiv:2609.37647}}}}.

\bibitem{Bradley97AUROC}
\bibinfo{author}{Bradley, A.~P.}
\newblock \bibinfo{title}{The use of the area under the {ROC} curve in the
  evaluation of machine learning algorithms}.
\newblock \emph{\bibinfo{journal}{Pattern Recognit.}}
  \textbf{\bibinfo{volume}{30}}, \bibinfo{pages}{1145--1159}
  (\bibinfo{year}{1997}).
\newblock \urlprefix\url{https://doi.org/10.1016/S0031-3203(96)00142-2}.

\bibitem{irvin2019Chexpert}
\bibinfo{author}{Irvin, J.} \emph{et~al.} \emph{\bibinfo{title}{Chexpert: {A}
  large chest radiograph dataset with uncertainty labels and expert
  comparison}}.
\newblock \emph{\bibinfo{booktitle}{The Thirty-Third {AAAI} Conference on
  Artificial Intelligence, {AAAI} 2019, The Thirty-First Innovative
  Applications of Artificial Intelligence Conference, {IAAI} 2019, The Ninth
  {AAAI} Symposium on Educational Advances in Artificial Intelligence, {EAAI}
  2019, Honolulu, Hawaii, USA, January 27 - February 1, 2019}},
  \bibinfo{pages}{590--597} (\bibinfo{publisher}{{AAAI} Press},
  \bibinfo{year}{2019}).
\newblock \urlprefix\url{https://doi.org/10.1609/aaai.v33i01.3301590}.

\bibitem{johnson2019mimicCxrJpg}
\bibinfo{author}{Johnson, A.} \emph{et~al.}
\newblock \bibinfo{title}{Mimic-cxr-jpg-chest radiographs with structured
  labels}.
\newblock \emph{\bibinfo{journal}{PhysioNet}} \textbf{\bibinfo{volume}{101}},
  \bibinfo{pages}{215--220} (\bibinfo{year}{2019}).

\bibitem{johnson2019mimic}
\bibinfo{author}{Johnson, A.~E.} \emph{et~al.}
\newblock \bibinfo{title}{Mimic-cxr, a de-identified publicly available
  database of chest radiographs with free-text reports}.
\newblock \emph{\bibinfo{journal}{Scientific data}}
  \textbf{\bibinfo{volume}{6}}, \bibinfo{pages}{317} (\bibinfo{year}{2019}).

\bibitem{goldberger2000physionet}
\bibinfo{author}{Goldberger, A.~L.} \emph{et~al.}
\newblock \bibinfo{title}{Physiobank, physiotoolkit, and physionet: components
  of a new research resource for complex physiologic signals}.
\newblock \emph{\bibinfo{journal}{circulation}} \textbf{\bibinfo{volume}{101}},
  \bibinfo{pages}{e215--e220} (\bibinfo{year}{2000}).

\bibitem{kather2018NctCrcHe100k}
\bibinfo{author}{Kather, J.~N.}, \bibinfo{author}{Halama, N.} \&
  \bibinfo{author}{Marx, A.}
\newblock \bibinfo{title}{100,000 histological images of human colorectal
  cancer and healthy tissue} (\bibinfo{year}{2018}).

\bibitem{bajwa2020g1020}
\bibinfo{author}{Bajwa, M.~N.} \emph{et~al.} \emph{\bibinfo{title}{{G1020:} {A}
  benchmark retinal fundus image dataset for computer-aided glaucoma
  detection}}.
\newblock \emph{\bibinfo{booktitle}{2020 International Joint Conference on
  Neural Networks, {IJCNN} 2020, Glasgow, United Kingdom, July 19-24, 2020}},
  \bibinfo{pages}{1--7} (\bibinfo{publisher}{{IEEE}}, \bibinfo{year}{2020}).
\newblock \urlprefix\url{https://doi.org/10.1109/IJCNN48605.2020.9207664}.

\bibitem{tschandl2018ham10000}
\bibinfo{author}{Tschandl, P.}, \bibinfo{author}{Rosendahl, C.} \&
  \bibinfo{author}{Kittler, H.}
\newblock \bibinfo{title}{The ham10000 dataset, a large collection of
  multi-source dermatoscopic images of common pigmented skin lesions}.
\newblock \emph{\bibinfo{journal}{Scientific data}}
  \textbf{\bibinfo{volume}{5}}, \bibinfo{pages}{1--9} (\bibinfo{year}{2018}).

\bibitem{yilmaz2024derm12345}
\bibinfo{author}{Yilmaz, A.}, \bibinfo{author}{Yasar, S.~P.},
  \bibinfo{author}{Gencoglan, G.} \& \bibinfo{author}{Temelkuran, B.}
\newblock \bibinfo{title}{Derm12345: A large, multisource dermatoscopic skin
  lesion dataset with 40 subclasses}.
\newblock \emph{\bibinfo{journal}{Scientific Data}}
  \textbf{\bibinfo{volume}{11}}, \bibinfo{pages}{1302} (\bibinfo{year}{2024}).

\end{thebibliography}
\end{document}